%% file: main.tex
\documentclass{article}

\usepackage{microtype}
\usepackage{graphicx}
\usepackage{booktabs}
\usepackage{wrapfig}
\usepackage{subcaption}
\usepackage{hyperref}
\usepackage{multirow}
\usepackage{xspace}

\usepackage{array}
\usepackage{amsthm}
\usepackage{amsmath}
\usepackage{colortbl}

\definecolor{Gray}{gray}{0.85}
\newcolumntype{g}{>{\columncolor{Gray}}c}

\input{math_commands.tex}

\newcommand*{\eg}{\textit{e.g.}\@\xspace}
\newcommand*{\ie}{\textit{i.e.}\@\xspace}

\newcommand{\indep}{\perp\!\!\!\perp}
\newcommand{\nindep}{\not\!\perp\!\!\!\perp}

\newcommand{\tr}{\text{tr}}
\newcommand{\te}{\text{te}}

\newcommand\methodname{\texttt{ReBias}\xspace}
\newcommand\arch[1]{\texttt{#1}}
\newcommand\method[1]{\texttt{#1}}

\usepackage{hyperref}

\usepackage[accepted]{icml2020}

\icmltitlerunning{Learning De-biased Representations with Biased Representations}

\begin{document}

\input{main_paper}

\input{supplementary_materials}

\end{document}

%% file: math_commands.tex
\usepackage{amsmath,amsfonts,bm}

\def\eqref#1{equation~\ref{#1}}

\def\1{\bm{1}}

\DeclareMathAlphabet{\mathsfit}{\encodingdefault}{\sfdefault}{m}{sl}
\SetMathAlphabet{\mathsfit}{bold}{\encodingdefault}{\sfdefault}{bx}{n}

\newcommand{\R}{\mathbb{R}}

\DeclareMathOperator*{\argmin}{arg\,min}

%% file: main_paper.tex
\twocolumn[
\icmltitle{Learning De-biased Representations with Biased Representations}

\begin{icmlauthorlist}
\icmlauthor{Hyojin Bahng}{kor}
\icmlauthor{Sanghyuk Chun}{naver}
\icmlauthor{Sangdoo Yun}{naver}
\icmlauthor{Jaegul Choo}{kaist}
\icmlauthor{Seong Joon Oh}{naver}
\end{icmlauthorlist}

\icmlaffiliation{kor}{Korea University}
\icmlaffiliation{naver}{Clova AI Research, NAVER Corp.}
\icmlaffiliation{kaist}{Graduate School of AI, KAIST}
\icmlcorrespondingauthor{Seong Joon Oh}{coallaoh@gmail.com}
\vskip 0.3in
]
\begin{NoHyper}
\printAffiliationsAndNotice{}
\end{NoHyper}

\begin{abstract}
Many machine learning algorithms are trained and evaluated by splitting data from a single source into training and test sets.
While such focus on \textit{in-distribution} learning scenarios has led to interesting advancement, it has not been able to tell if models are relying on dataset biases as shortcuts for successful prediction (\eg, using snow cues for recognising snowmobiles), resulting in biased models that fail to generalise when the bias shifts to a different class. 
The \textit{cross-bias generalisation} problem has been addressed by de-biasing training data through augmentation or re-sampling, which are often prohibitive due to the data collection cost (\eg, collecting images of a snowmobile on a desert) and the difficulty of quantifying or expressing biases in the first place. 
In this work, we propose a novel framework to train a de-biased representation by encouraging it to be \textit{different} from a set of representations that are biased by design. 
This tactic is feasible in many scenarios where it is much easier to define a set of biased representations than to define and quantify bias. 
We demonstrate the efficacy of our method across a variety of synthetic and real-world biases; our experiments show that the method discourages models from taking bias shortcuts, resulting in improved generalisation. 
Source code is available at \url{https://github.com/clovaai/rebias}.
\end{abstract}

\section{Introduction}
\label{sec:intro}
Most machine learning algorithms are trained and evaluated by randomly splitting a single source of data into training and test sets. Although this is a standard protocol, it is blind to a critical problem: the reliance on dataset bias~\citep{torralba2011unbiased}. For instance, many frog images are taken in swamp scenes, but swamp itself is not a frog. Nonetheless, a model will exploit this bias (\ie, take ``shortcuts'') if it yields correct predictions for the majority of training examples. If the bias is sufficient to achieve high accuracy, there is little motivation for models to learn the complexity of the intended task, despite its full capacity to do so. Consequently, a model that relies on bias will achieve high in-distribution accuracy, yet fail to generalise when the bias shifts.

We tackle this ``cross-bias generalisation'' problem where a model does not exploit its full capacity due to the ``sufficiency'' of bias cues for prediction of the target label in the training data. For example, language models make predictions based on the presence of certain words (\eg, ``not'' for ``contradiction'')~\citep{gururangan2018naacl} without much reasoning on the actual meaning of sentences, even if they are in principle capable of sophisticated reasoning. Similarly, convolutional neural networks (CNNs) achieve high accuracy on image classification by using local texture cues as shortcut, as opposed to more reliable global shape cues~\citep{StylisedImageNet, BagNet}. 3D CNNs achieve high accuracy on video action recognition by relying on static cues as shortcut rather than capturing temporal actions~\citep{weinzaepfel2019mimetics, RESOUND, REPAIR}. 

Existing methods attempt to remove a model's dependency on bias by de-biasing the training data through augmentation~\citep{StylisedImageNet} or introducing a pre-defined bias that a model is trained to be independent of~\citep{HEX}. Other approaches~\citep{clark2019don,cadene2019rubi} learn a biased model given source of bias as input, and de-bias through logit re-weighting or logit ensembling. These prior studies assume that biases can be easily defined or quantified (\ie, explicit bias label), but often real-world biases do not (\eg, texture or static bias above).

To address this limitation, we propose a novel framework to train a de-biased representation by encouraging it to be statistically independent from representations that are biased by design. We use the Hilbert-Schmidt Independence Criterion~\citep{HSIC} to formulate the independence. Our insight is that there are certain types of bias that can be easily captured by defining a bias-characterising model (\eg, CNNs of smaller receptive fields for texture bias; 2D CNNs for static bias in videos). Experiments show that our method effectively reduces a model's dependency on ``shortcuts'' in training data; accuracy is improved in test data where the bias is shifted or removed.

\begin{figure*}[ht!]
	\centering
	\includegraphics[width=0.8\linewidth]{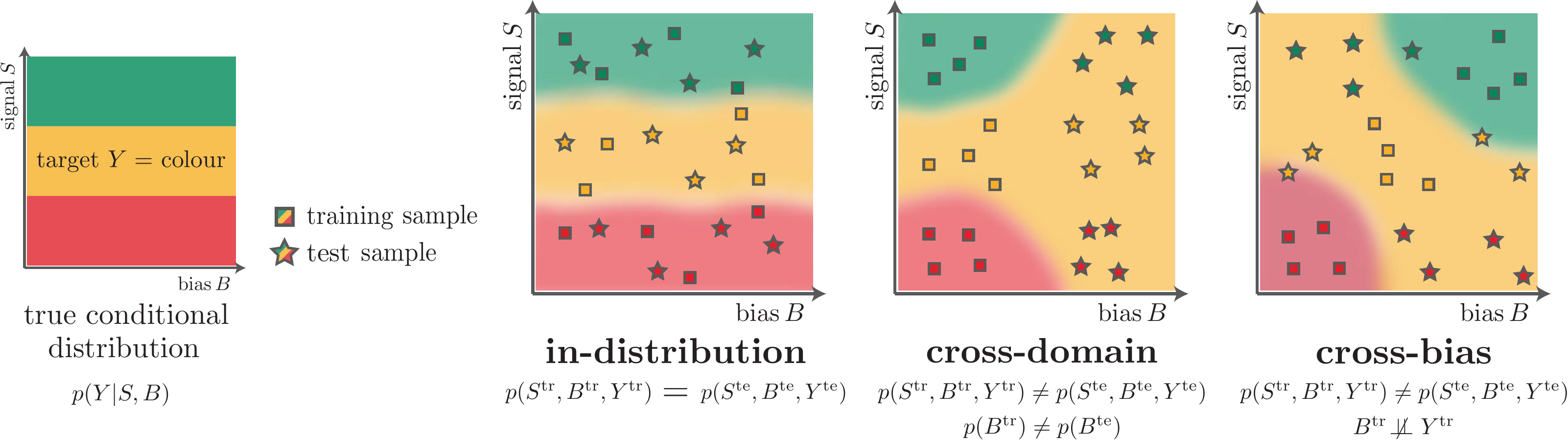}
	\vspace{-.8em}
	\caption{\small \textbf{Learning scenarios}.
	Different distributional gaps may take place between training and test distributions. Our work addresses the \textit{cross-bias generalisation} problem. Background colours on the right three figures indicate the decision boundaries of models trained on given training data.}
	\label{fig:learning-scenarios}
\end{figure*}

\section{Problem Definition}
\label{sec:problem}

We provide a rigorous definition of our over-arching goal: overcoming the bias in models trained on biased data. We systematically categorise the learning scenarios and cross-bias generalisation strategies.

\subsection{Cross-bias generalisation}
\label{subsec:problem-generalisation}

We first define random variables, signal $S$ and bias $B$ as cues for the recognition of an input $X$ as certain target variable $Y$. Signals $S$ are the cues essential for the recognition of $X$ as $Y$; examples include the shape and skin patterns of frogs for frog image classification. Biases $B$'s, on the other hand, are cues not essential for the recognition but correlated with the target $Y$; many frog images are taken in swamp scenes, so swamp scenes can be considered as $B$. A key property of $B$ is that intervening on $B$ should not change $Y$; moving a frog from swamp to a desert scene does not change the ``frogness''. We assume that the true predictive distribution $p(Y|X)$ factorises as $\int p(Y|S,B) p(S,B|X)$, signifying the sufficiency of $p(S,B|X)$ for recognition. 

Under this framework, three learning scenarios are identified depending on the change of relationship $p(S,B,Y)$ across training and test distributions,  $p(S^{tr},B^{tr},Y^{tr})$ and $p(S^{te},B^{te},Y^{te})$, respectively: in-distribution, cross-domain, and cross-bias generalisation. See Figure~\ref{fig:learning-scenarios}.

\paragraph{In-distribution.}
$p(S^{tr},B^{tr},Y^{tr})=p(S^{te},B^{te},Y^{te})$. This is the standard learning setup utilised in many benchmarks by splitting data from a single source into training and test data at random.

\paragraph{Cross-domain.}
$p(S^{tr},B^{tr},Y^{tr})\neq p(S^{te},B^{te},Y^{te})$ and furthermore $p(B^{tr})\neq p(B^{te})$. $B$ in this case is often referred to as ``domain''. For example, training data consist of images with ($Y^{tr}$=frog, $B^{tr}$=wilderness) and ($Y^{tr}$=bird, $B^{tr}$=wilderness), while test data contain ($Y^{te}$=frog, $B^{te}$=indoors) and ($Y^{te}$=bird, $B^{te}$=indoors). This scenario is typically simulated by training and testing on different datasets~\citep{ben2007analysis}.

\paragraph{Cross-bias.}
$p(B^{tr}) \nindep p(Y^{tr})$\footnote{$\indep$ and $\nindep$ denote independence and dependence, respectively.} and the dependency changes across training and test distributions: $p(B^{tr},Y^{tr}) \neq p(B^{te},Y^{te})$. We further assume that $p(B^{tr})= p(B^{te})$, to clearly distinguish the scenario from the cross-domain generalisation. For example, training data only contain images of two types ($Y^{tr}$=frog, $B^{tr}$=swamp) and ($Y^{tr}$=bird, $B^{tr}$=sky), but test data contain unusual class-bias combinations ($Y^{te}$=frog, $B^{te}$=sky) and ($Y^{te}$=bird, $B^{te}$=swamp). Our work addresses this scenario.

\subsection{Existing cross-bias generalisation methods and their assumptions}
\label{subsec:problem-prior-work}

Under cross-bias generalisation scenarios, the dependency $p(B^{tr}) \nindep p(Y^{tr})$ makes bias $B$ a viable cue for recognition. The model trained on such data becomes susceptible to interventions on $B$, limiting its generalisabililty when the bias is changed or removed in the test data. There exist prior approaches to this problem, but with different types and amounts of assumptions on $B$. We briefly recap the approaches based on the assumptions they require. In the next part \S\ref{subsec:problem-function-class}, we will define our problem setting that requires an assumption distinct from the ones in prior approaches.

\paragraph{When an algorithm to disentangle bias $B$ and signal $S$ exists.}

Being able to disentangle $B$ and $S$ lets one collapse the feature space corresponding to $B$ in both training and test data. A model trained on such normalised data then becomes free of biases. As ideal as it is, building a model to disentangle $B$ and $S$ is often unrealistic (\eg, texture bias~\citep{StylisedImageNet}). Thus, researchers have proposed other approaches to tackle cross-bias generalisation.

\paragraph{When a data collection procedure or generative algorithm for $p(X|B)$ exists.} 

When additional examples can be supplied through $p(X|B)$, the training dataset itself can be de-biased, \ie, $B\indep Y$. Such a data augmentation strategy is indeed a valid solution adopted by many prior studies. Some approach has proposed to collect additional data to balance out the bias~\citep{panda2018eccv}. Other approaches have proposed to synthesise data with a generative algorithm through image stylisation~\citep{StylisedImageNet}, object removal~\citep{agarwal2019towards,shetty2019cvpr}, or generation of diverse, semantically similar linguistic variations~\citep{shah2019cycle,ray2019sunny}. However, collecting unusual inputs can be expensive~\citep{peyre2017iccv}, and building a generative model with pre-defined bias types~\citep{StylisedImageNet} may suffer from bias mis-specification or the lack of realism.

\paragraph{When a ground truth or predictive algorithm for $p(B|X)$ exists.}

Conversely, when one can tell the bias $B$ for every input $X$, we can remove the dependency between the model predictions $f(X)$ and the bias $B$. The knowledge on $p(B|X)$ is provided in many realistic scenarios. For example, when the aim is to remove gender biases $B$ in a job application process $p(Y|X)$, applicants' genders $p(B|X)$ are supplied as ground truths. Many existing approaches for fairness in machine learning have proposed independence-based regularisers to encourage $f(X)\indep B$~\citep{zemel2013icml} or the conditional independence $f(X)\indep B\,|\, Y$~\citep{quadrianto2019discovering, Separation}. Other approaches have proposed to remove predictability of $p(B|X)$ based on $f(X)$ through domain adversarial losses~\citep{louppe2017learning,wang2019balanced} or mutual information minimisation~\citep{kim2019learning,creager2019flexibly}. When the ground truth of $p(B|X)$ is not provided, another approach has proposed to quantify texture bias by utilising the neural gray-level co-occurrence matrix and encouraging independence through projection~\citep{HEX}. Unfortunately, for certain bias types (\eg, texture bias), it is difficult to enumerate the possible bias classes and put labels on samples. 

\subsection{Our scenario: Capturing bias with a set of models}
\label{subsec:problem-function-class}

Under the cross-bias generalisation scenario, some biases are not easily addressed by the above methods. Take texture bias as an example (\S\ref{sec:intro}, \citet{StylisedImageNet}): (1) texture $B$ and shape $S$ cannot easily be disentangled, (2) collecting unusual images or building a generative model $p(X|B)$ is expensive, (3) building the predictive model $p(B|X)$ for texture requires enumeration (classifier) or embedding (regression) of all possible textures, which is not feasible. 

However, slightly modifying the third assumption results in a problem setting that allows interesting application scenarios. Instead of assuming explicit knowledge on $p(B|X)$, we can approximate $B$ by defining a set of models $G$ that are biased towards $B$ by design. For texture biases, for example, we define $G$ to be the set of CNN architectures with small receptive fields. Then, any learned model $g\in G$ can by design make predictions $g(x)$ based on the patterns that can only be captured with small receptive fields (\ie, textures), becoming more liable to overfit to texture.

More precisely, we define $G$ to be a \textbf{bias-characterising model class} for the bias-signal pair $(B,S)$ if for every possible joint distribution $p(B,X)$ there exists a $g\in G$ such that $p(B|X)\approx g(X)$ (\textbf{recall condition}) and every $g\in G$ satisfies $g(X) \indep S \,|\, B$ (\textbf{precision condition}). Consider these conditions as conceptual tools to break down “what is a good $G$?" In practice, $G$ may not necessarily include all biases and may also capture important signals (\ie, imperfect recall and precision). With this in mind, we formulate our framework as a regulariser to the original task so that $f(X)$ does not ignore every signal captured by $G$. We do not require $G$ to be perfect.

There exist many scenarios when such $G$ can be characterised, based on several empirical evidence for the type of bias. For instance, action recognition models rely heavily on static cues without learning temporal cues~\citep{RESOUND,REPAIR,choi2019can}; we can regularise the 3D CNNs towards better generalisation across static biases by defining G to be the set of 2D CNNs. VQA models rely overly on language biases rather than visual cues~\citep{agrawal2018don}. G can be defined as the set of models that only look at the language modality~\citep{clark2019don, cadene2019rubi}. Entailment models are biased towards word overlap rather than understanding the underlying meaning of sentences~\citep{mccoy2019right,niven2019probing}. We can design G to be the set of bag-of-words classifiers~\citep{clark2019don}. These scenarios exemplify situations when the added architectural capacity is not fully utilised because there exist simpler cues for solving the task.

There are recent approaches that attempt to capture bias with bias-characterising models $G$ and remove dependency on $B$ via logit ensembling~\citep{clark2019don} or logit re-weighting~\citep{cadene2019rubi}. In \S\ref{sec:experiments}, we empirically measure their performance on synthetic and realistic biases.

\section{Proposed Method}
\label{sec:method}

We present a solution for the cross-bias generalisation when the bias-characterising model class $G$ is known (see \S\ref{subsec:problem-function-class}); the method is referred to as \methodname. The solution consists of training a model $f$ for the task $p(Y|X)$ with a regularisation term encouraging the independence between the prediction $f(X)$ and the set of all possible biased predictions $\{g(X)\,|\,g\in G\}$. We will introduce the precise definition of the regularisation term and discuss why and how it leads to the unbiased model.

\subsection{\methodname: Removing bias with bias}
\label{subsec:method}

If $p(B|X)$ is fully known, we can directly encourage $f(X)\indep B$. Since we only have access to the set of biased models $G$ (\S\ref{subsec:problem-function-class}), we seek to promote $f(X)\indep g(X)$ for every $g\in G$. Simply put, we de-bias a representation $f\in F$ by designing a set of biased models $G$ and letting $f$ run away from $G$. This leads to the independence from bias cues $B$ while leaving signal cues $S$ as valid recognition cues; see \S\ref{subsec:problem-function-class}. We will specify \methodname learning objective after introducing our independence criterion, HSIC.

\paragraph{Hilbert-Schmidt Independence Criterion (HSIC).}
Since we need to measure the degree of independence between continuous random variables $f(X)$ and $g(X)$ in high-dimensional spaces, it is infeasible to resort to histogram-based measures; we use HSIC~\citep{HSIC}. For two random variables $U$ and $V$ and kernels $k$ and $l$, HSIC is defined as $\text{HSIC}^{k,l}(U,V):=||C^{k,l}_{UV}||_{\text{HS}}^2$ where $C^{k,l}$ is the cross-covariance operator in the Reproducing Kernel Hilbert Spaces (RKHS) of $k$ and $l$~\citep{HSIC}, an RKHS analogue of covariance matrices. $||\cdot ||_{\text{HS}}$ is the Hilbert-Schmidt norm, a Hilbert-space analogue of the Frobenius norm. It is known that for two random variables $U$ and $V$ and radial basis function (RBF) kernels $k$ and $l$, $\text{HSIC}^{k,l}(U,V)=0$ if and only if $U\indep V$. A finite-sample estimate of $\text{HSIC}^{k,l}(U,V)$ has been used in practice for statistical testing~\citep{HSIC,HSICTest}, feature similarity measurement~\citep{kornblith2019similarity}, and model regularisation~\citep{quadrianto2019discovering,zhang2018ijcai}. We employ an unbiased estimator $\text{HSIC}^{k,l}_1(U,V)$~\citep{unbiasedHSIC} with $m$ samples, defined as
\begin{align}
\scriptstyle
    \text{HSIC}^{k,l}_1(U,V) =\frac{1}{m(m-3)}\left[\text{tr}(\widetilde{U}\widetilde{V}^T)     + \frac{\mathbf{1}^T \widetilde{U} \mathbf{1} \mathbf{1}^T \widetilde{V}^T \mathbf{1}}{(m-1)(m-2)} - \frac{2}{m-2} \mathbf{1}^T \widetilde{U} \widetilde{V}^T  \mathbf{1} \right] \nonumber
    \label{eq:unbiased_hsic}
\end{align}
where $\widetilde{U}_{ij}=(1-\delta_{ij})\,k(u_i,u_j)$, $\{u_i\}\sim U$, \ie, the diagonal entries of $\widetilde{U}$ are set to zero. $\widetilde{V}$ is defined similarly.

\paragraph{Minimax optimisation for bias removal.}
We define 
\begin{align}
  \text{HSIC}^k_1(f(X),G(X)):=\underset{g\in G}{\max}\,\,\text{HSIC}^k_1(f(X),g(X))
\end{align}
with an RBF kernel $k$ for the degree of independence between representation $f\in F$ and the biased representations $G$. We write $\text{HSIC}_1(f,G)$ and $\text{HSIC}_1(f,g)$ as shorthands. The learning objective for $f$ is then defined as 
\begin{align}
  \underset{f\in F}{\min}\,\,\left\{\mathcal{L}(f,X,Y) + \lambda\,\,\underset{g\in G}{\max}\,\,\text{HSIC}_1(f,g)\right\}
  \label{eq:main_prev}
\end{align}
where $\mathcal{L}(f,X,Y)$ is the loss for the main task $p(Y|X)$ and $\lambda>0$. We write $\mathcal{L}(f)$ as shorthands. Having specified $G$ to represent the bias $B$, we need to train $g\in G$ for the original task to intentionally overfit $G$ to $B$. Thus, the inner optimisation involves both the independence criterion and the original task loss $\mathcal{L}(g)$. The final learning objective for \methodname is then
\begin{align}
  \underset{f}{\min}\,\left\{\mathcal{L}(f) + \lambda\,\underset{g}{\max} \Big( \text{HSIC}_1(f,g) - \lambda_{g} \mathcal{L}(g) \Big) \right\}.
  \label{eq:main}
\end{align}
$\lambda,\lambda_{g} > 0$. We solve \eqref{eq:main} by alternative updates. Our intention is that $f$ is trained to be different from multiple possible biased predictions $\{g(X)\,|\,g\in G\}$, thereby improving its de-biased performance.

\subsection{Why and how does it work?}
\label{subsec:method-why-and-how}

Independence describes relationships between random variables, but we use it for function pairs. Which functional relationship does statistical independence translate to? In this part, we argue with proofs and observations that the answer to the above question is \textit{the dissimilarity of invariance types learned by a pair of models}.

\paragraph{Linear case: Equivalence between independence and orthogonality.}

We study the set of function pairs $(f,g)$ satisfying $f(X)\indep g(X)$ for suitable random variable $X\sim p(X)$. Assuming linearity of involved functions and the normality of $X$, we obtain the equivalence between statistical independence and functional orthogonality. 

\paragraph{Lemma 1.}
Assume that $f$ and $g$ are affine mappings $f(x)=Ax+a$ and $g(x)=Bx+b$ where $A\in\R^{m\times n}$ and $B\in\R^{l\times n}$. Assume further that $X$ is a normal distribution with mean $\mu$ and covariance matrix $\Sigma$. Then, $f(X)\indep g(X)$ if and only if $\ker(A)^\perp\perp_{\Sigma}\ker(B)^\perp$. For a positive semi-definite matrix $\Sigma$, we define $\langle r, s\rangle_\Sigma = \langle r, \Sigma s \rangle$, and the set orthogonality $\perp_{\Sigma}$ likewise. The proof is in Appendix.

In particular, when $f$ and $g$ have 1-dimensional outputs, the independence condition is translated to the orthogonality of their weight vectors and decision boundaries; $f$ and $g$ are models with orthogonal invariance types.

\paragraph{Non-linear case: HSIC as a metric learning objective.}

We lack theories to fully characterise general, possibly non-linear, function pairs $(f,g)$ achieving $f(X)\indep g(X)$; it is an interesting open question. For now, we make a set of observations in this general case, using the finite-sample independence criterion $\text{HSIC}_0(f,g):=(m-1)^{-2}\text{tr}(\widetilde{f}\,\,\widetilde{g}^T)=0$, where $\widetilde{f}$ is the mean-subtracted kernel matrix $\widetilde{f}_{ij}=k(f(x_i),f(x_j))-m^{-1} \sum_{k} k(f(x_i),f(x_k))$ and likewise for $\widetilde{g}$. Unlike in the loss formulation (\S\ref{subsec:method}), we use the biased HSIC statistic for simplicity.

Note that $\text{tr}(\widetilde{f}\,\,\widetilde{g}^T)$ is an inner product between flattened matrices $\widetilde{f}$ and $\widetilde{g}$. We consider the inner-product-minimising solution for $f$ on an input pair $x_0 \neq x_1$ given a fixed $g$. The problem can be written as $\min_{f(x_0),f(x_1)}\,\,\text{tr}(\widetilde{f}\,\,\widetilde{g}^T)$, which is equivalent to $\min_{f(x_0),f(x_1)}\,\,\widetilde{f}_{01}\cdot\widetilde{g}_{10}$.

When $\widetilde{g}_{10}>0$, $g$ is relatively invariant on $(x_1,x_0)$, since $k(g(x_1),g(x_0))>m^{-1} \sum_{i} k(g(x_1),g(x_i))$. Then, the above problem boils down to $\min_{f(x_0),f(x_1)}\,\,\widetilde{f}_{01}$, signifying the relative \textit{variance} of $f$ on $(x_0,x_1)$. Following a similar argument, we obtain the converse statement: if $g$ is relatively variant on a pair of inputs, invariance of $f$ on the pair minimises the objective.

We conclude that $\min_f\,\,\text{HSIC}_0(f,g)$ against a fixed $g$ is a metric-learning objective for the embedding $f$, where ground truth pairwise matches and mismatches are relative mismatches and matches for $g$, respectively. As a result, $f$ and $g$ learn different sorts of invariances.

\paragraph{Effect of HSIC regularisation on toy data.}
We have established that HSIC regularisation encourages the difference in model invariances. To see how it helps to de-bias a model, we have prepared synthetic two-dimensional training data following the cross-domain generalisation case in Figure~\ref{fig:learning-scenarios}: $X=(B,S)\in \R^2$ and $Y\in\{\text{red, yellow, green}\}$. Since the training data is perfectly biased, a multi-layer perceptron (MLP) trained on the data only shows 55\% accuracy on de-biased test data (see decision boundary figure in Appendix). To overcome the bias, we have trained another MLP with \eqref{eq:main} where the bias-characterising class $G$ is defined as the set of MLPs that take only the bias dimension as input. This model exhibits de-biased decision boundaries (Appendix) with improved accuracy of 89\% on the de-biased test data.

\section{Experiments}
\label{sec:experiments}

In the previous section, \methodname has been introduced and theoretically justified. In this section, we present experimental results of \methodname. We first introduce the setup, including the biases tackled in the experiments, difficulties inherent to the cross-bias evaluation, and the implementation details (\S\ref{subsec:experimental-setup}). Results on Biased MNIST (\S\ref{subsec:mnist}), ImageNet (\S\ref{subsec:imagenet}) and action recognition (\S\ref{subsec:action}) are shown afterwards. While our experiments are focused on vision tasks, we stress that the underlying concept and methodology are not exclusive to them. For example, it will be an interesting future research direction to apply \methodname to visual question answering and natural language understanding problems (\S\ref{subsec:problem-function-class}), tasks that also suffer from dataset biases.

\subsection{Experimental setup}
\label{subsec:experimental-setup}

\paragraph{Which biases do we tackle?}

Our work tackles the types of biases that are used as shortcut cues for recognition in the training data. In the experiments, we tackle the ``texture'' bias in image classification and the ``static'' bias in video action recognition. Even if a CNN image classifier has wide receptive fields, empirical evidence indicates that they heavily rely on local texture cues for recognition, instead of the global shape cues~\citep{StylisedImageNet}. Similarly, a 3D CNN action recognition model possesses the capacity to model temporal cues, yet it heavily relies on static cues like scenes or objects rather than the temporal motion for recognition~\citep{weinzaepfel2019mimetics}. While it is difficult to precisely define and quantify all texture or scene types, it is easy to intentionally design a model $G$ biased towards such cues. In other words, we model the entire bias domain (\eg, texture) through a chosen inductive bias of the $G$ network architecture. For texture bias in image recognition, we design $G$ as a CNN with smaller receptive fields; for static bias in action recognition, we design $G$ as a 2D CNN. 
\paragraph{Evaluating cross-bias generalisation is difficult.}
To measure the performance of a model across real-world biases, one requires an unbiased dataset or one where the types and degrees of biases can be controlled. Unfortunately, data in real world arise with biases. To de-bias a frog and bird image dataset with swamp and sky (see \S\ref{subsec:problem-generalisation}), either rare data samples must be collected or one must generate such data; they are expensive procedures~\citep{peyre2017iccv}.

We thus evaluate our method along two axes: (1) synthetic biases (Biased MNIST) and (2) realistic biases (ImageNet classification and action recognition task). Biased MNIST contains colour biases which we control in training and test data for an in-depth analysis of \methodname. For ImageNet classification, on the other hand, we use clustering-based proxy ground truths for texture bias to measure the cross-bias generalisability. For action recognition, we utilize the unbiased data that are publicly available (Mimetics), albeit in small quantity. We use the Mimetics dataset~\citep{weinzaepfel2019mimetics} for the unbiased test set accuracies, while using the biased Kinetics~\citep{carreira2017quo} dataset for training. The set of experiments complement each other in terms of experimental control and realism.

\begin{table*}[ht!]
\centering
\setlength\tabcolsep{0.4em}
\small
\begin{tabular}{cccccccgccccccg}
 & & \multicolumn{6}{c}{Biased} & & \multicolumn{6}{c}{Unbiased} \\
\cline{3-8} \cline{10-15}
& \vspace{-1em} \\
$\rho$ & & {\scriptsize Vanilla} & {\scriptsize Biased} & {\scriptsize\method{HEX}} & {\scriptsize\method{LearnedMixin}} & {\scriptsize\method{RUBi}} & {\scriptsize\methodname (ours)} & & {\scriptsize Vanilla} & {\scriptsize Biased} & {\scriptsize\method{HEX}} & {\scriptsize\method{LearnedMixin}} & {\scriptsize\method{RUBi}} & {\scriptsize\methodname (ours)} \\
\cline{1-1} \cline{3-8} \cline{10-15}
& \vspace{-1em} \\
\cline{1-1} \cline{3-8} \cline{10-15}
& \vspace{-1em} \\
.999             &  & \textbf{100.}           & \textbf{100.}        & 71.3 & 2.9        & 99.9  & \textbf{100.} &  & 10.4            & 10.  &      10.8       & 12.1       & 13.7 & \textbf{22.7} \\
.997             &  & \textbf{100.}           & \textbf{100.}        & 77.7  & 6.7        & 99.4  & \textbf{100.} &  & 33.4            & 10.     & 16.6        & 50.2       & 43.0 & \textbf{64.2} \\
.995             &  & \textbf{100.}           & \textbf{100.}         & 80.8 & 17.5       & 99.5  & \textbf{100.} &  & 72.1            & 10.     & 19.7         & 78.2       & \textbf{90.4} & 76.0 \\
.990          &  & \textbf{100.}           & \textbf{100.}      & 66.6   & 33.6       & \textbf{100.} & \textbf{100.} &  & 89.1         & 10.   & 24.7           & 88.3       & \textbf{93.6} & 88.1 \\
\cline{1-1} \cline{3-8} \cline{10-15}
& \vspace{-1em} \\
avg.           &  & \textbf{100.}           & \textbf{100.}     & 74.1     & 15.2       & 99.7  & \textbf{100.} &  & 51.2            & 10.    & 18.0          & 57.2       & 60.2 & \textbf{62.7} \\
\cline{1-1} \cline{3-8} \cline{10-15}
& \vspace{-1em} \\
\end{tabular}
\caption{\small\textbf{Biased MNIST results.} Biased and unbiased accuracies on varying train correlation $\rho$. Besides our results, we report vanilla $F$, $G$ and previous methods. Ours are shown in gray columns. Each value is the average of three different runs.}
\label{table:mnist-experiment}
\end{table*}

\paragraph{Implementation of \methodname.}

We describe the specific design choices in \methodname implementation (\eqref{eq:main}). The source code is in the supplementary materials.

For texture biases, we define the biased model architecture families $G$ as CNNs with small receptive fields (RFs). The biased models in $G$ will by design learn to predict the target class of an image only through the local texture cues. On the other hand, we define a larger search space $F$ with larger RFs for our unbiased representations.

In our work, all networks $f$ and $g$ are fully convolutional networks followed by a global average pooling (GAP) layer and a linear classifier. $f(x)$ and $g(x)$ denote the outputs of GAP layer (feature maps), on which we compute the independence measures using HSIC (\S\ref{subsec:method}).

For Biased MNIST, $F$ is a fully convolutional network with four convolutional layers with $7\times 7$ kernels. Each convolutional layer uses batch normalisation~\citep{batchnorm} and ReLU. $G$ has the same architecture as $F$, except that the kernel sizes are $1\times 1$. On ImageNet, we use the \arch{ResNet18}~\citep{ResNet} architecture for $F$ with RF$=$435. $G$ is defined as \arch{BagNet18}~\citep{BagNet}, which replaces many $3\times 3$ kernels with $1\times 1$, thereby being limited to RF$=$43. For action recognition, we use \arch{3D-ResNet18} and \arch{2D-ResNet18} for $F$ and $G$ whose RF along temporal dimension are 19 and 1, respectively.

We conduct experiments using the same batch size, learning rate, and epochs for fair comparison. We choose $\lambda = \lambda_g = 1$. For the Biased MNIST experiments, we set the kernel radius to one, while the median of distances is chosen for ImageNet and action recognition experiments. More implementation details are provided in Appendix.

\vspace{-.8em}
\paragraph{Comparison methods.}
There are prior methodologies that can be applied in our cross-bias generalisation task (\S\ref{subsec:problem-prior-work}). We empirically compare \methodname against them. The prior methods include \method{RUBi}~\citep{cadene2019rubi} and \method{LearnedMixin+H}~\citep{clark2019don} that reduce the dependency of the model $F$ on biases captured by $G$ via logit re-weighting and logit ensembling, respectively. While the prior works additionally alter the training data for $G$, we only compare the objective functions themselves in our experiment. We additionally compare two methods that tackle texture bias: \method{HEX}~\citep{HEX} and \method{StylisedImageNet}~\citep{StylisedImageNet}. \method{HEX} attempts to reduce the dependency of a model on ``superficial statistics''. It measures texture via neural grey-level co-occurrence matrices (NGLCM) and projects out the NGLCM feature from the model. \method{StylisedImageNet} reduces the model’s reliance on texture by augmenting the training data with texturised images.

\subsection{Biased MNIST}
\label{subsec:mnist}

We first verify our model on a dataset where we have full control over the type and amount of bias during training and evaluation. We describe the dataset and present the experimental results.

\subsubsection{Dataset and evaluation}
\label{subsubsec:mnist-dataset} 

We construct a new dataset called \textbf{Biased MNIST} designed to measure the extent to which models generalise to bias shift. We modify MNIST~\citep{lecun1998gradient} by introducing the colour bias that highly correlate with the label $Y$ during training. With $B$ alone, a CNN can achieve high accuracy without having to learn inherent signals for digit recognition $S$, such as shape, providing little motivation for the model to learn beyond these superficial cues. 

\begin{figure}[h!]
\centering
\includegraphics[width=\linewidth]{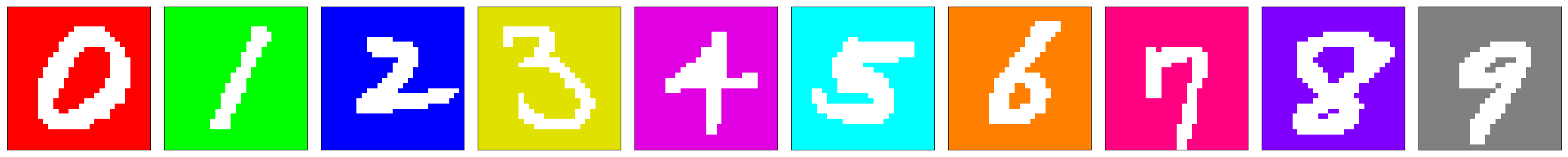}
\vspace{-2em}
\caption{\small\textbf{Biased MNIST.} A synthetic dataset with the colour bias which highly correlates with the labels during training.}
\label{fig:biased_mnist}
\end{figure}

We inject the colour bias by adding a colour on training image backgrounds (Figure~\ref{fig:biased_mnist}). We pre-select 10 distinct colours for each digit $y\in\{0,\cdots,9\}$. Then, for each image of digit $y$, we assign the pre-defined colour $b(y)$ with probability $\rho\in [0,1]$ and any other colour with probability $(1-\rho)$. $\rho$ then controls the bias-target correlation in the training data: $\rho=1.0$ leads to complete bias and $\rho=0.1$ leads to an unbiased dataset. We consider $\rho \in \{0.99, 0.995, 0.997, 0.999\}$ to simulate significant amounts of bias during training. We evaluate the model’s generalisability to bias shift by evaluating under the following criterion:

\paragraph{Biased.} $p(S^\te,B^\te,Y^\te)=p(S^\tr,B^\tr,Y^\tr)$, the in-distribution case in \S\ref{subsec:problem-generalisation}. Whatever bias the training set contains, it is replicated in the test set (same $\rho$). This measures the ability of de-biased models to maintain high in-distribution performances while generalising to the cross-bias test set. 

\paragraph{Unbiased.}
$B^\te\indep Y^\te$, the cross-bias generalization in \S\ref{subsec:problem-generalisation}. We assign biases on test images independently of the labels. Bias is no longer predictive of $Y$ and a model needs to utilise actual signals $S$ to yield correct predictions.

\begin{figure*}
\centering
\small
\begin{tabular}{cccccc}
    \includegraphics[width=.17\linewidth]{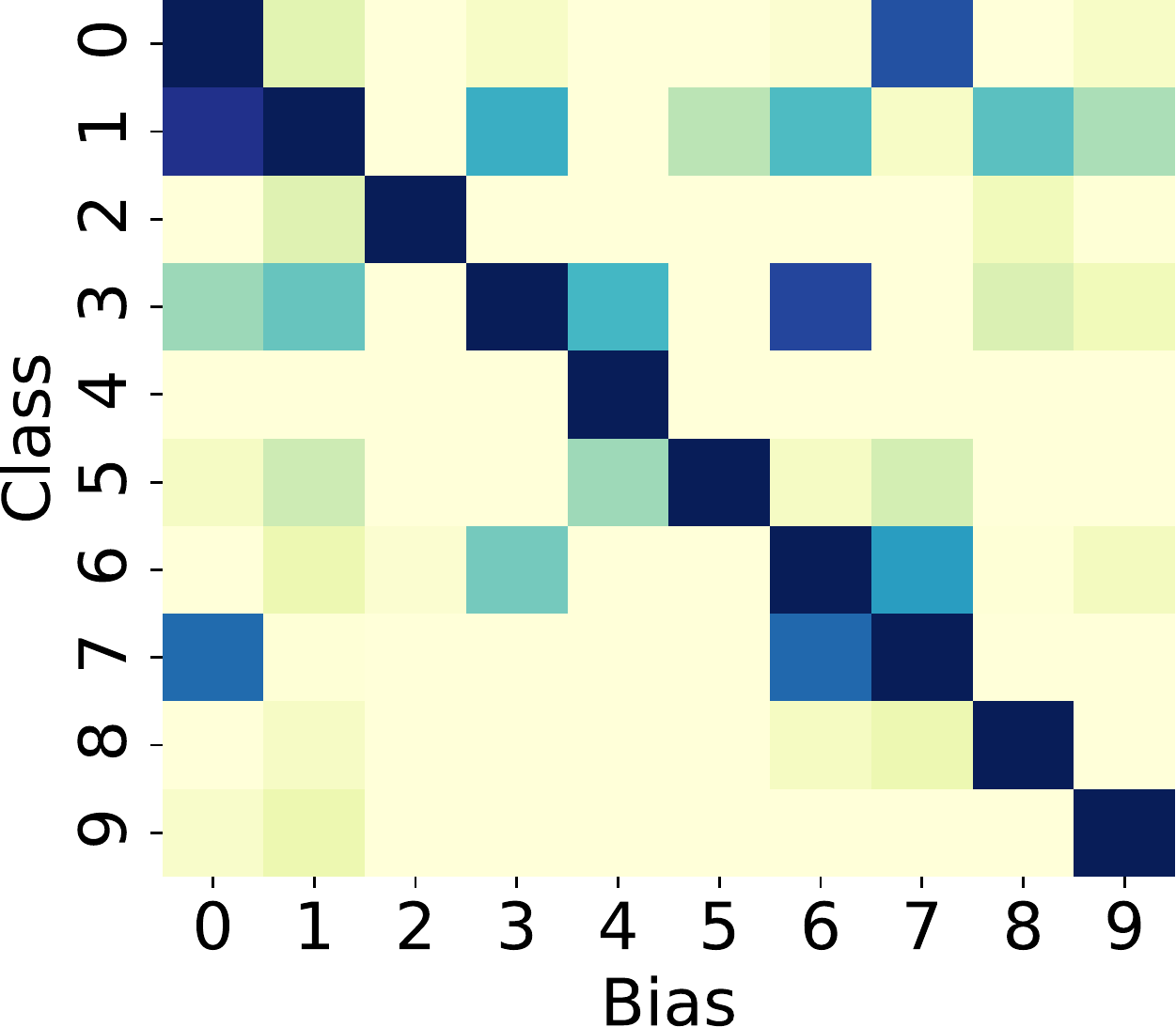} & 
    \includegraphics[width=.17\linewidth]{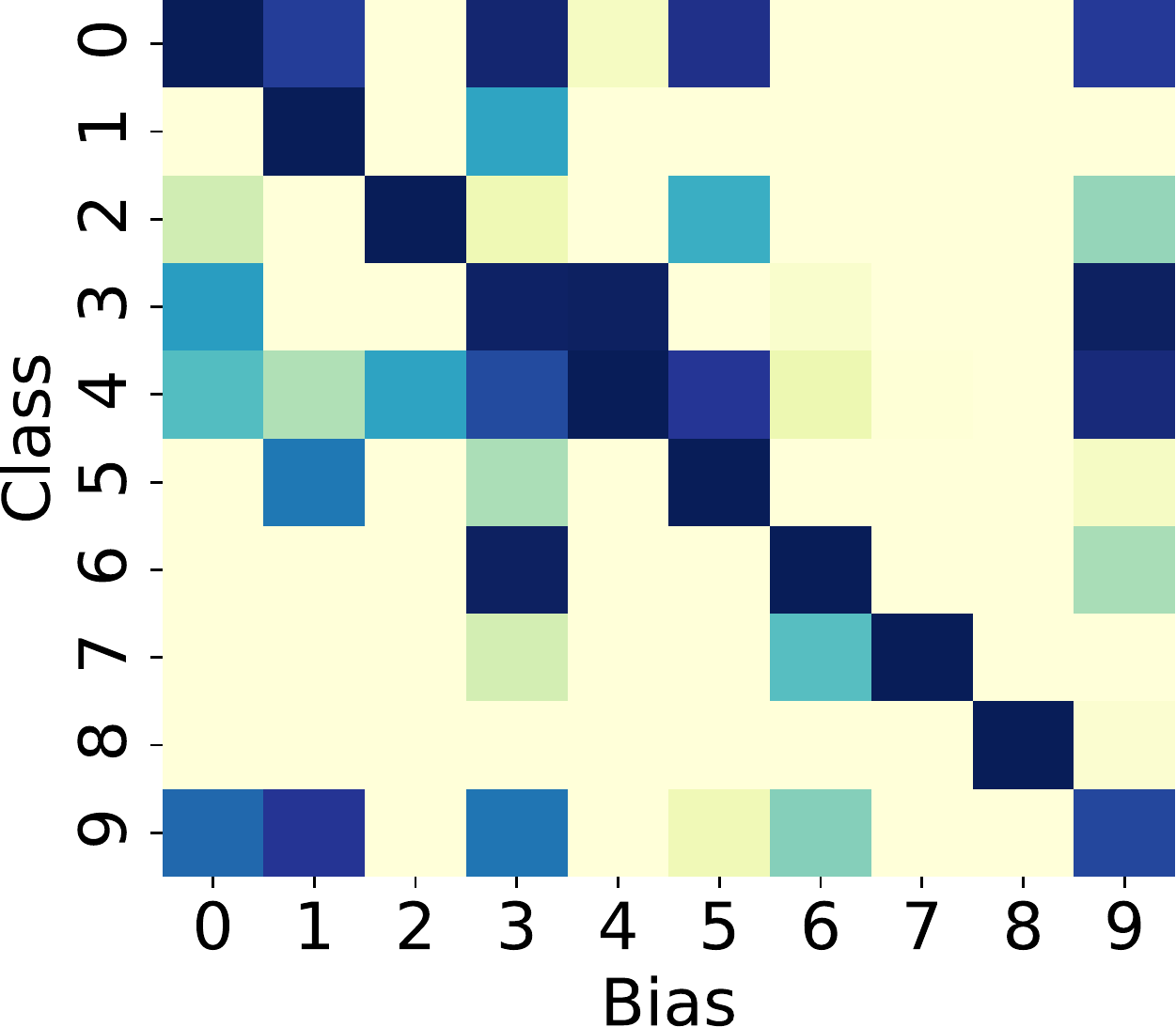} & 
    \includegraphics[width=.17\linewidth]{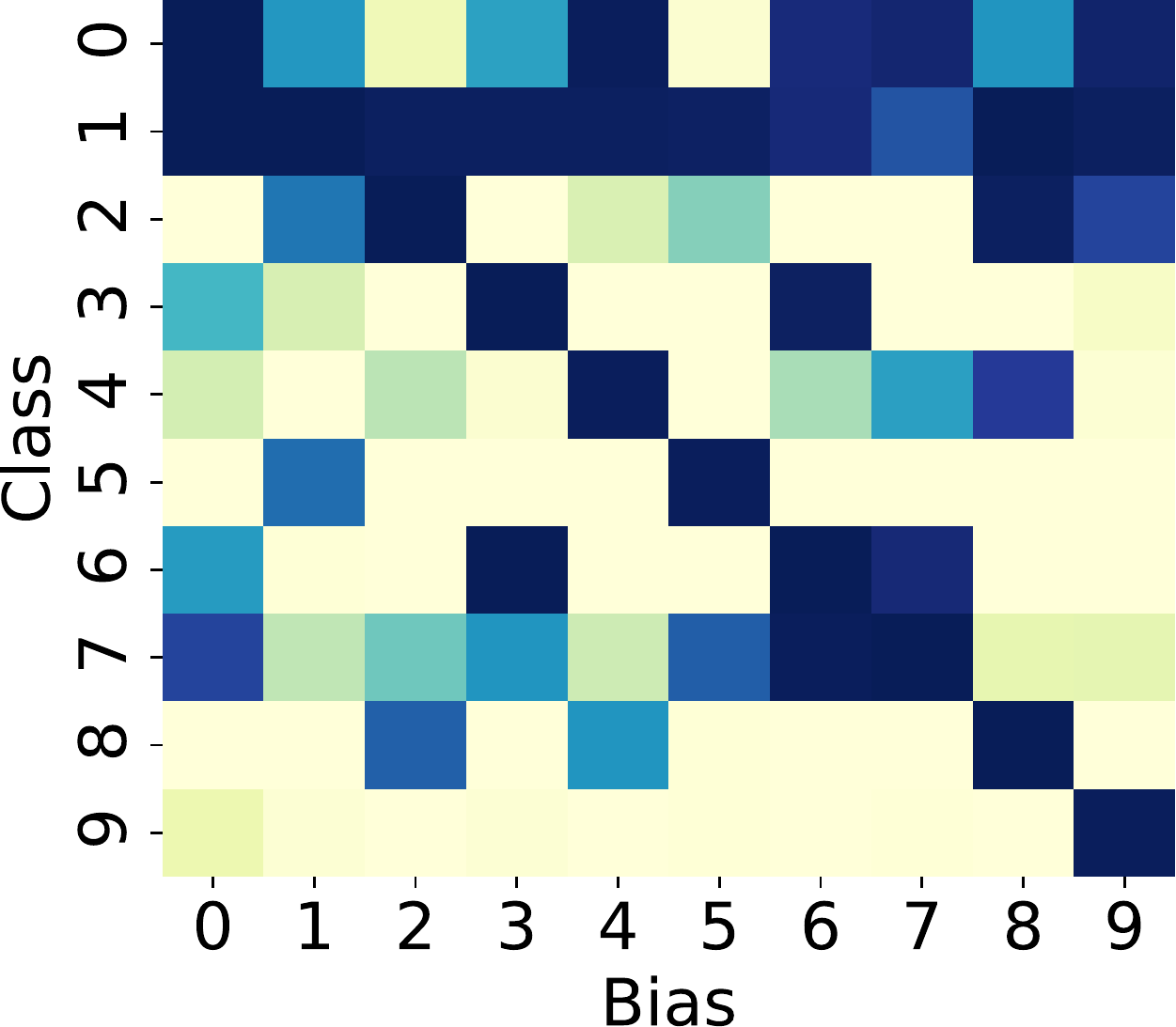} & 
    \includegraphics[width=.17\linewidth]{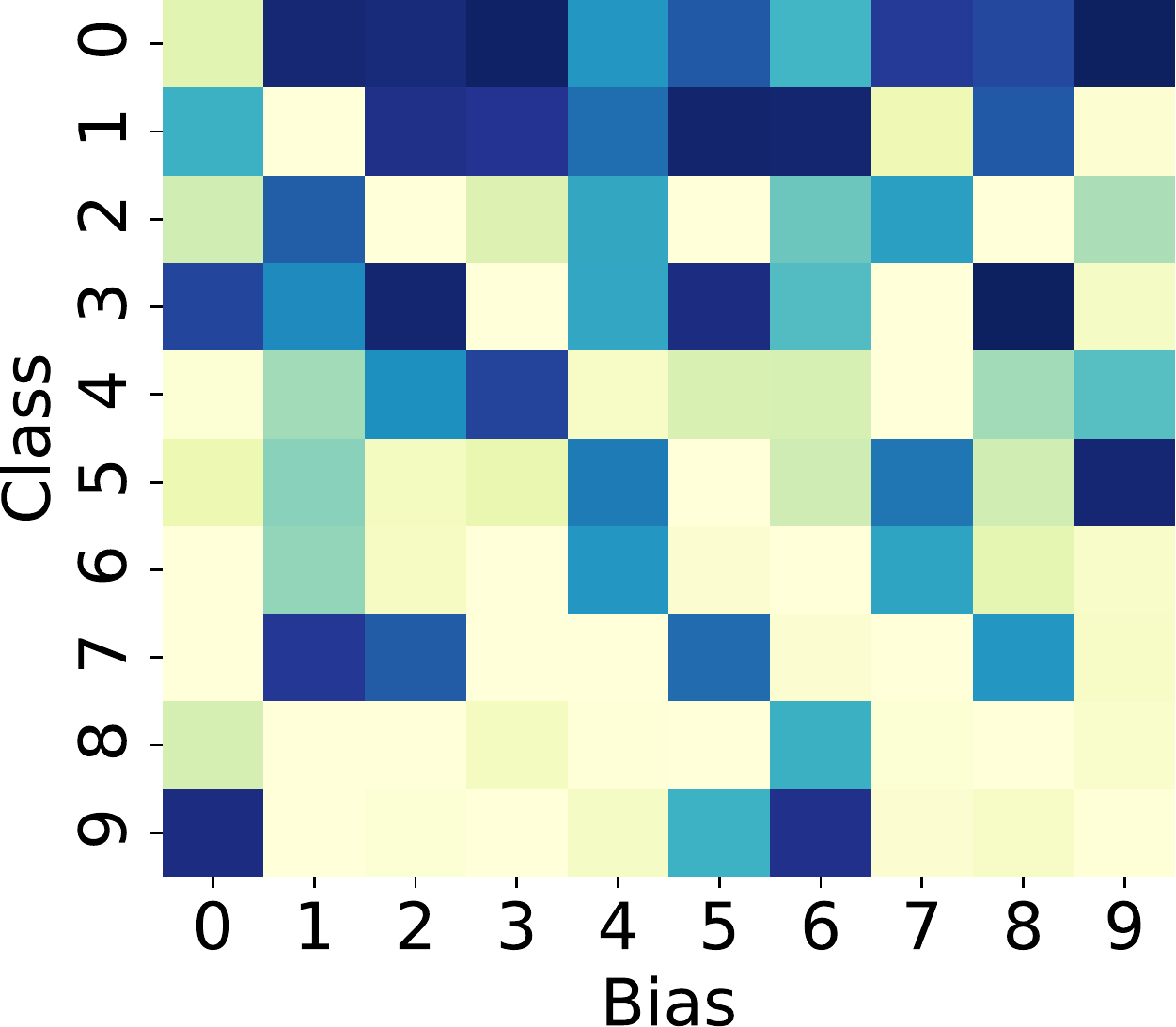} & 
    \includegraphics[width=.17\linewidth]{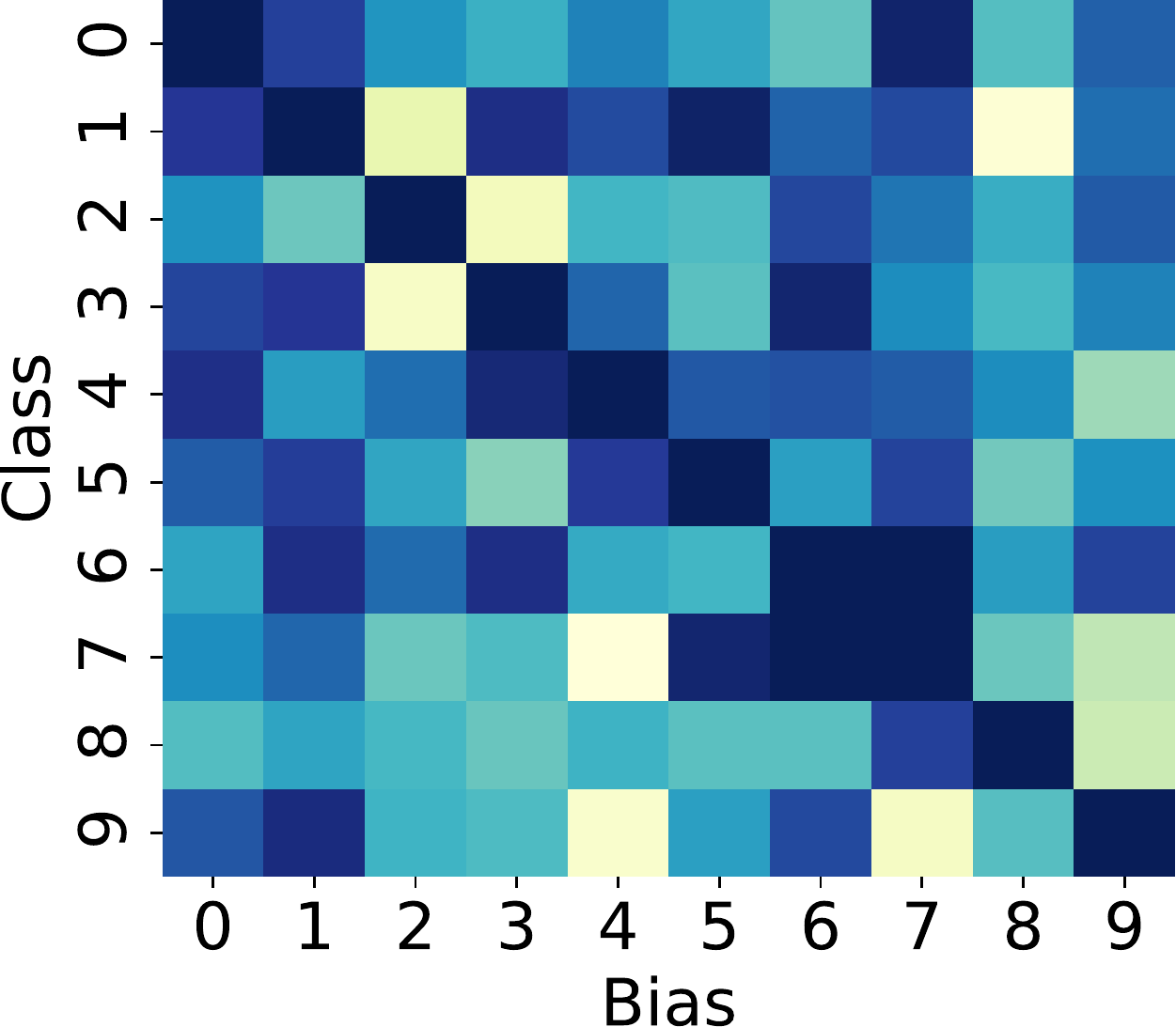} & 
    \includegraphics[width=.03\linewidth]{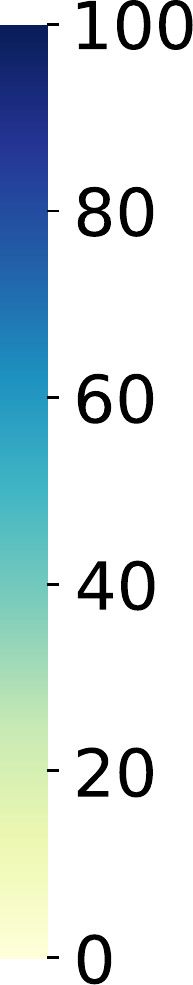} \\
    Vanilla & \method{HEX} & \method{RUBi} & \method{LeanredMixin} & \methodname
\end{tabular}
\caption{\small \textbf{Accuracy per bias-class pair.} We show accuracies for each bias and class pair $(B,Y)=(b,y)$ on Biased MNIST. All methods are trained with $\rho=0.997$. The diagonals in each matrix indicate the pre-defined bias-target pair (\S\ref{subsubsec:mnist-dataset}). The number of samples per $(b,y)$ cell is identical across all pairs (unbiased test set).}
\label{fig:mnist_cellwise_accuracies}
\end{figure*}

\subsubsection{Results}
\label{subsubsec:mnist-results}

Results on the Biased MNIST are shown in Table~\ref{table:mnist-experiment}.

\paragraph{\methodname lets a model overcome bias.}

We observe that vanilla $F$ achieves 100\% accuracy under the ``biased'' metric (the same bias between training and test data) in the Biased MNIST for all $\rho$. This is how most machine learning tasks are evaluated, yet this does not show the extent to which the model depends on bias for prediction. When the bias cues are randomly assigned to the label at evaluation, vanilla $F$ accuracy collapses to 10.4\% under the ``unbiased'' metric on the Biased MNIST when the train correlation is large, \ie, $\rho=0.999$. The intentionally biased models $G$ result in 10.0\% on the Biased MNIST, the random chance performance, for all $\rho$. This exemplifies the case where a seemingly high-performing model has in fact overfitted to bias and does not generalise to new situations.

On the other hand, \methodname achieves robust generalisation across all settings by learning to be different from the representations $G$. Especially, \methodname unbiased accuracies than the vanilla model under the highly correlated settings: $10.4\rightarrow22.7\%$ and $33.4\rightarrow 64.2\%$ boosts for $\rho=$0.999 and 0.997, respectively.

\paragraph{Comparison against other methods.}
As \method{HEX} pre-defines bias as patterns captured by NGLCM, we observe that it does not improve generalisability to colour bias (18.0\%) while also hurting the in-distribution accuracy (74.1\%) compared to vanilla $F$. \method{LearnedMixin} achieves performance gain in unbiased accuracies (57.2\%) yet suffers a severe performance drop for unbiased accuracies (15.2\%). \method{RUBi} achieves robust generalisation across biased and unbiased accuracies (99.7\% and 60.2\% respectively). We show in the following experiments that \method{LearnedMixin} and \method{RUBi} achieve sub-optimal performances in realistic texture and static biases. 

\paragraph{Analysis of per-bias performances.}
In Figure~\ref{fig:mnist_cellwise_accuracies}, we provide more fine-grained results by visualising the accuracies per bias-class pair $(B,Y)=(b,y)$. The diagonal average corresponds to the \textit{biased accuracy} and the overall average corresponds to the \textit{unbiased accuracy}. We observe that the vanilla model has higher accuracies on diagonals and lower on off-diagonals, showing the heavy reliance on colour (bias) cues. \method{HEX} and \method{RUBi} demonstrate sporadic improvements in certain off-diagonals, but the overall improvements are limited. \method{LearnedMixin} shows further enhancements, yet with near-zero accuracies on diagonal entries (also seen in Table~\ref{table:mnist-experiment}). \methodname uniformly improves the off-diagonals, while not sacrificing the diagonals.

\begin{figure}[h]
\centering
\includegraphics[width=.8\linewidth]{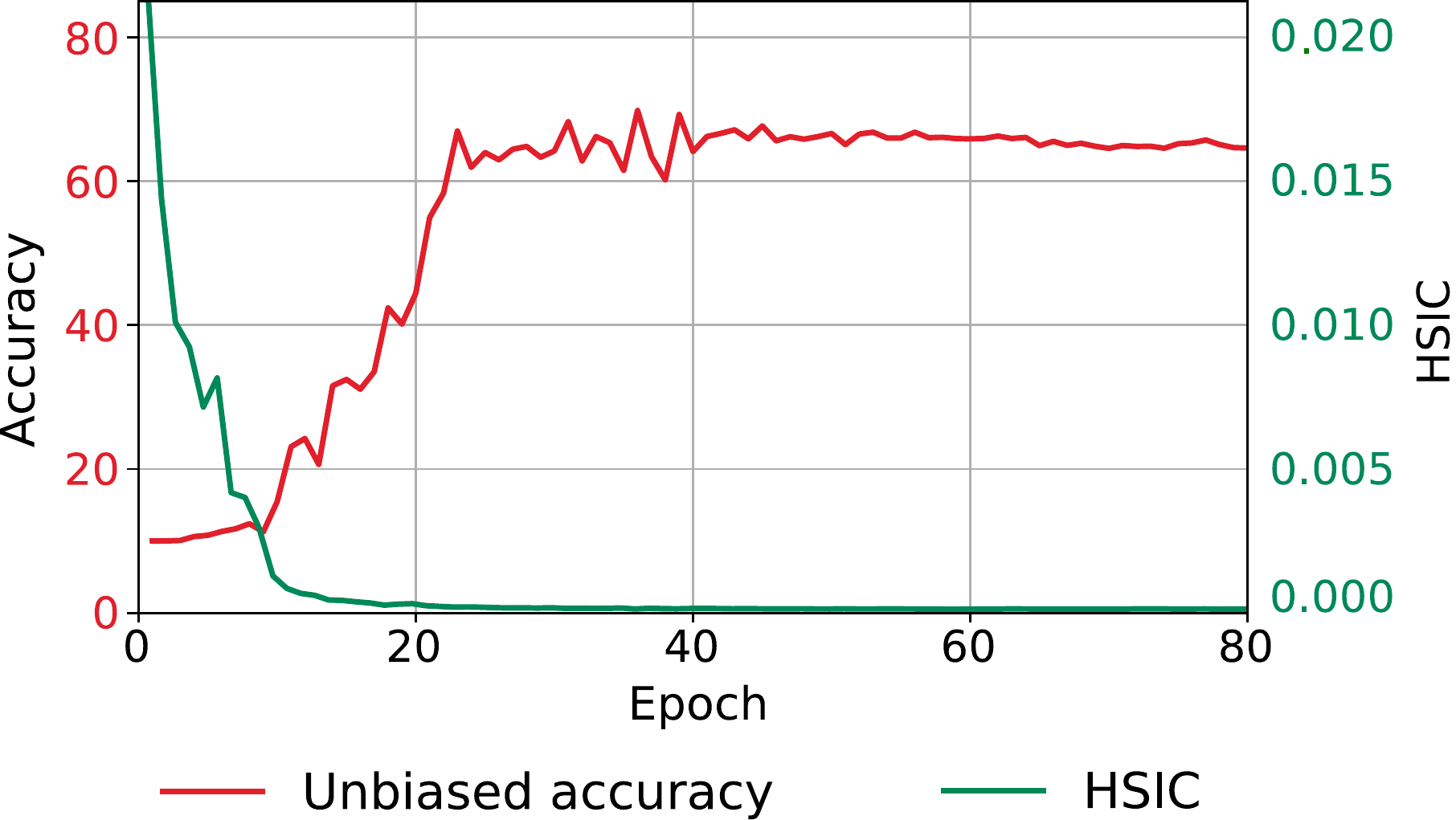}
\vspace{-.5em}
\caption{\small \textbf{Learning curves.} \methodname achieves better generalisation by minimizing HSIC between representations.}
\label{fig:mnist-learning-curve}
\end{figure}

\paragraph{Learning Curves.}
In Figure~\ref{fig:mnist-learning-curve}, we plot the evolution of unbiased accuracy and HSIC values as \methodname is trained. \methodname is trained with $\rho=0.997$ and tested with $\rho=0.1$ (unbiased). 
While the classification loss alone, \ie, vanilla $F$, leads to an unbiased accuracy of $33.4\%$, the unbiased accuracy increases dramatically ($>60\%$) as the HSIC between $F$ and $G$ is minimised during training. We observe that there exists a strong correlation between the HSIC values and unbiased accuracies.

\subsection{ImageNet}
\label{subsec:imagenet}
In ImageNet experiments, we further validate the applicability of \methodname on the texture bias in realistic images (\ie, objects in natural scenes). The texture bias often lets a model achieve good in-distribution performances by exploiting the local texture shortcuts (\eg, determining a swan class by not seeing its shape but the background water texture).

\subsubsection{Dataset and evaluation}
\label{subsubsec:imagenet-dataset}

We construct \textbf{9-Class ImageNet}, a subset of ImageNet~\citep{russakovsky2015imagenet} containing 9 super-classes ~\citep{ilyas2019adversarial}, since using the original ImageNet is not scalable. We additionally balance the ratios of sub-class images for each super-class to focus on the effect of texture bias. 

Since it is difficult to evaluate the cross-bias generalisability on realistic data (\S\ref{subsec:experimental-setup}), we settle for surrogate measures:
\paragraph{Biased.} 
$p(S^\te,B^\te,Y^\te)=p(S^\tr,B^\tr,Y^\tr)$. Accuracy is measured on the in-distribution validation set. Though widely-used, this metric is blind to a model's generalisability to unseen bias-target combinations. 
\paragraph{Unbiased.} 
$B^\te\indep Y^\te$.
As a proxy to the perfectly de-biased test data, which is difficult to collect (\S\ref{subsec:experimental-setup}), we use \emph{texture} clusters IDs $c\in\{1,\cdots,K\}$ as the ground truth labels for texture bias obtained by the k-means clustering. For full details of texture clustering algorithm, see Appendix. For an unbiased accuracy measurement, we compute the accuracies for every set of images corresponding to a texture-class combination $(c,y)$. The combination-wise accuracy $A_{c,y}$ is computed by $\text{Corr}(c,y)/\text{Pop}(c,y)$, where $\text{Corr}(c,y)$ is the number of correctly predicted samples in $(c,y)$ and $\text{Pop}(c,y)$ is the total number of samples in $(c,y)$, called the \textbf{population} at $(c,y)$. The unbiased accuracy is then the mean accuracy over all $A_{c,y}$ where the population $\text{Pop}(c,y)>10$. This measure gives more weights on samples of unusual texture-class combinations (smaller $\text{Pop}(c,y)$) that are less represented in the usual biased accuracies. Under this unbiased metric, a biased model basing its recognition on textures is likely to show sub-optimal results on unusual combinations, leading to a drop in the unbiased accuracy. Since the k-means clustering is non-convex, we report the average unbiased accuracy of three clustering results with different initial points.

\paragraph{ImageNet-A.} ImageNet-A~\citep{hendrycks2019natural} contains the failure cases of ImageNet-trained \arch{ResNet50} among web images. The images consist of many failure modes of networks when ``frequently appearing background elements''~\citep{hendrycks2019natural} become erroneous cues for recognition (\eg a bee image feeding on hummingbird feeder is recognised as a hummingbird). An improved performance on ImageNet-A is an indirect signal that the model learns beyond the bias shortcuts.

\paragraph{ImageNet-C.} 
ImageNet-C~\citep{hendrycks2018imagenet-c} is proposed to evaluate robustness to $15$ corruption types including ``noise'', ``blur'', ``weather'', and ``digital'' with five severities. Improved performances on ImageNet-C indicate that the model robustly generalises to a wide range of distortions, despite their absence during training.

\subsubsection{Results}
\label{subsubsec:imagenet-results}

We measure performances of \arch{ResNet18} trained under \methodname to be different from \arch{BagNet18}. We use the metrics in the previous part. Results are shown in Table~\ref{table:imagenet-experiment}. 

\paragraph{Vanilla models are biased.}
\arch{ResNet18} shows good performances on the biased accuracy (90.8\%) but dropped performances on the texture-unbiased accuracy (88.8\%). \arch{BagNet18} performs worse than the vanilla ResNet as they are heavily biased towards texture by design (\ie, small receptive field sizes). The drop signifies the biases of vanilla models towards texture cues; by basing their predictions on texture cues they obtain generally better accuracies on texture-class pairs $(c,y)$ that are more represented. The drop also shows the limitation of current evaluation schemes where the cross-bias generalisation is not measured.

\begin{table}[t]
\centering
\setlength\tabcolsep{0.4em}
\small
\begin{tabular}{lccccc}
Model description && Biased  & Unbiased & IN-A & IN-C \\ \cline{1-1} \cline{3-6}
\vspace{-1em} & \\ 
\cline{1-1} \cline{3-6}
\vspace{-1em} & \\
Vanilla {\scriptsize(\arch{ResNet18})} && 90.8      &  88.8    & 24.9 & 54.2 \\  
Biased {\scriptsize(\arch{BagNet18})} && 67.7      &  65.9    & 18.8 & 31.7 \\   
\cline{1-1} \cline{3-6}
\vspace{-1em} & \\
{\scriptsize\method{StylisedIN}} {\tiny\cite{StylisedImageNet}} && 88.4  & 86.6 & 24.6 & \textbf{61.1} \\
{\scriptsize\method{LearnedMixin}} {\tiny\cite{clark2019don}} && 64.1 & 62.7 & 15.0 & 27.5 \\ 
{\scriptsize\method{RUBi}} {\tiny\cite{cadene2019rubi}} && 90.5 & 88.6 & 27.7 & 53.7 \\
\cline{1-1} \cline{3-6}
\vspace{-1em} & \\
\cellcolor{Gray}\methodname (ours) && \cellcolor{Gray}\textbf{91.9} & \cellcolor{Gray}\textbf{90.5} & \cellcolor{Gray}\textbf{29.6} & \cellcolor{Gray}{57.5}\\
\cline{1-1} \cline{3-6}
\end{tabular}
\caption{\small\textbf{ImageNet results.} We show results corresponding to $F=$\arch{ResNet18} and $G=$\arch{BagNet18}. IN-A and IN-C indicates ImageNet-A and ImageNet-C, respectively. We repeat each experiment three times.}
\label{table:imagenet-experiment}
\end{table}

\paragraph{\methodname leads to less biased models.}
When \methodname is applied on \arch{ResNet18} to make it learn cues beyond those captured by \arch{BagNet18}, we observe a general boost in the biased, unbiased, ImageNet-A, and ImageNet-C accuracies (Table~\ref{table:imagenet-experiment}). The unbiased accuracy of \arch{ResNet18} improves from 88.8\% to 90.5\%, thus robustly generalising to less represented texture-class combinations at test time. Our method also shows improvements on the challenging ImageNet-A subset (\eg from 24.9\% to 29.6\%), which further shows an improved generalisation. While \method{StylisedImageNet} attempts to mitigate texture bias by stylisation, it does not increase the generalisability for both the unbiased and ImageNet-A accuracy (86.6\% and 24.6\% respectively). Similar to the Biased MNIST results, \method{Learned-Mixin} suffers a collapse in the in-distribution accuracy (from 90.8\% to 67.9\%) and does not improve generalisability to less represented texture-class combinations or the challenging ImageNet-A. \method{RUBi} only shows improvement on ImageNet-A (from 24.9\% to 27.7\%).

In ImageNet-C experiments, \method{StylisedImageNet} shows the best ImageNet-C performance (61.1\%) and \methodname achieves the second best accuracy (57.5\%). Despite \methodname does not use any data augmentation, it improves the generalisability to ImageNet-C while \method{Learned-Mixin} and \method{RUBi} fail to generalise (from 54.2\% to 27.5\% and 53.7\%, respectively).

\subsection{Action recognition}
\label{subsec:action}
To see further effectiveness of \methodname on reducing static biases in a video understanding task, we conduct the action recognition experiments with 3D CNNs. 3D CNNs have proven their state-of-the-art performances on action recognition benchmarks such as Kinetics~\citep{carreira2017quo}, but recent studies~\citep{sevilla2019onlytimecantell, RESOUND, REPAIR} have shown that such action datasets have strong static biases towards the scene or objects in videos. As a result, 3D CNNs make predictions dominantly based on static cues, despite their ability to capture temporal signals, and they achieve high accuracies even with temporal cues removed (\eg, shuffling frames or masking-out human actor in videos)~\citep{weinzaepfel2019mimetics}. This bias problem occasionally leads to performance drop when static cues shift across training and test settings (\eg, predicting ``swimming'' class when a person plays football near a swimming pool).

\subsubsection{Dataset}
We use the Kinetics dataset~\citep{carreira2017quo} for training, which is known to have bias towards static cues. To evaluate the cross-bias generalisability, we use the Mimetics dataset~\citep{weinzaepfel2019mimetics} that consists of videos of a mime artist performing actions without any context. The classes of Mimetics are fully covered by the Kinetics classes and we use it as the unbiased validation set. Since the training and testing of the full action datasets are not scalable, we sub-sample 10-classes from both datasets. Detailed dataset descriptions are in Appendix.

\subsubsection{Results}
We evaluate the performances of \arch{3D-ResNet18} trained to be different from the biased model \arch{2D-ResNet18}.
Main results are shown in Table~\ref{table:action-experiment}.

\paragraph{Vanilla model is biased.}
The vanilla \arch{3D-ResNet18} model $F$ shows a reasonable performance on the biased Kinetics with 54.5\% accuracy, but significantly loses the accuracy on the unbiased Mimetics with 18.9\% accuracy. While \arch{3D-ResNet18} is originally designed for capturing temporal signals within videos, it relies a lot on static cues, resulting in a similar performance with the 18.4\% accuracy by \arch{2D-ResNet18} on Mimetics.

\begin{table}[t]
\centering
\setlength\tabcolsep{0.4em}
\small
\begin{tabular}{lcccc}
&& Biased     & Unbiased   \\
Model description && (Kinetics) & (Mimetics) \\
\cline{1-1} \cline{3-4}
\vspace{-1em} & \\ 
\cline{1-1} \cline{3-4}
\vspace{-1em} & \\
Vanilla {\scriptsize(\arch{3D-ResNet18})} &&  54.5 &  18.9    \\  
Biased {\scriptsize(\arch{2D-ResNet18})} &&  50.7 &  18.4    \\  
\cline{1-1} \cline{3-4}
& \vspace{-1em} \\
{\scriptsize\method{LearnedMixin}} {\scriptsize\cite{clark2019don}} && 12.3 & 11.4 \\
{\scriptsize\method{RUBi}} {\scriptsize\cite{cadene2019rubi}}  && 22.4  & 13.4 \\
\cline{1-1} \cline{3-4}
& \vspace{-1em} \\
\cellcolor{Gray} \methodname (ours) && \cellcolor{Gray}\textbf{55.8} & \cellcolor{Gray}\textbf{22.4} \\ 
\cline{1-1} \cline{3-4}
\end{tabular}
\caption{\textbf{Action recognition results.} We show results corresponding to $F=$\arch{3D-ResNet18} and $G=$\arch{2D-ResNet18} with baseline comparisons. Top-1 accuracies are reported. Each result is the average of three runs.}
\label{table:action-experiment}
\vspace{-1em}
\end{table}

\paragraph{\methodname reduces the static bias.}
Applying \methodname on \arch{3D-ResNet18} encourages it to utilise the temporal modelling capacity by forcing it to reason differently from \arch{2D-ResNet18}. \methodname improves the accuracies on both Kinetics and Mimetics datasets beyond the vanilla model $F$: $54.5\rightarrow55.8\%$ and $18.9\rightarrow 22.4\%$, respectively. We also compare against the two baseline methods, \method{LearnedMixin}~\citep{clark2019don} and \method{RUBi}~\citep{cadene2019rubi}, as in the previous sections. \methodname shows better performances than the two baseline methods for reducing the static bias for action recognition. We believe that the difficulty of the action recognition task on Kinetics hampers the normal operation of the logit-modification step in the baseline methods, severely hindering the convergence with respect to the cross-entropy loss. The training of \methodname, on the other hand, remains stable as the independence loss acts only as a regularisation term.

\section{Conclusion}
\label{sec:conclusion}

We have identified a practical problem faced by many machine learning algorithms that the learned models exploit bias shortcuts to recognise the target: the cross-bias generalisation problem (\S\ref{sec:problem}). Models tend to under-utilise its capacity to extract non-bias signals (\eg, global shapes for object recognition, or temporal actions for action recognition) when bias shortcuts provide sufficient cues for recognition in the training data (\eg, texture for object recognition, or static contexts for action recognition)~\citep{StylisedImageNet, weinzaepfel2019mimetics}. We have addressed this problem with the \methodname method. Given an identified set of models $G$ that encodes the bias to be removed, \methodname encourages a model $f$ to be statistically independent of $G$ (\S\ref{sec:method}). We have provided theoretical justifications (\S\ref{subsec:method-why-and-how}) and have validated the superiority of \methodname in removing biases from models through experiments on Biased MNIST, ImageNet classification, and the Mimetics action recognition benchmark (\S\ref{sec:experiments}). 

\section*{Acknowledgements}
We thank Clova AI Research team for the discussion and advice, especially Dongyoon Han, Youngjung Uh, Yunjey Choi, Byeongho Heo, Junsuk Choe, Muhammad Ferjad Naeem, and Hyojin Park for their internal reviews. Naver Smart Machine Learning (NSML) platform~\citep{kim2018nsml} has been used in the experiments. Kay Choi has helped the design of Figure~\ref{fig:learning-scenarios}. This work was partially supported by Institute of Information \& communications Technology Planning \& Evaluation (IITP) grant funded by the Korea government(MSIT) (No.2019-0-00075, Artificial Intelligence Graduate School Program(KAIST)) and the National Research Foundation of Korea (NRF) grant funded by the Korean government (MSIP) (No. NRF-2019R1A2C4070420).

{
\bibliography{references}
\bibliographystyle{icml2020}
}

%% file: supplementary_materials.tex
\onecolumn
\pdfstringdefDisableCommands{\def\\{}}
\icmltitle{Learning De-biased Representations with Biased Representations\\-- Appendix --}
\appendix

\setcounter{equation}{0}
\setcounter{figure}{0}
\setcounter{table}{0}

\renewcommand{\theequation}{A\arabic{equation}}
\renewcommand{\thefigure}{A\arabic{figure}}
\renewcommand{\thetable}{A\arabic{table}}

\section{Statistical Independence is Equivalent to Functional Orthogonality for Linear Maps}
\label{appendix:orthogonality-proof}

We provide a proof for the following lemma in \S3.2.

\paragraph{Lemma 1.}
Assume that $f$ and $g$ are affine mappings $f(x)=Ax+a$ and $g(x)=Bx+b$ where $A\in\R^{m\times n}$ and $B\in\R^{l\times n}$. Assume further that $X$ is a normal distribution with mean $\mu$ and covariance matrix $\Sigma$. Then, $f(X)\indep g(X)$ if and only if $\ker(A)^\perp\perp_{\Sigma}\ker(B)^\perp$. $\indep$ denotes the independence. For a positive semi-definite matrix $\Sigma$, we define $\langle r, s\rangle_\Sigma = \langle r, \Sigma s \rangle$. The orthogonality $V\perp_{\Sigma}W$ of two subspaces $V$ and $W$ is defined likewise.

\begin{proof}
Due to linearity and normality, the independence $f(X)\indep g(X)$ is equivalent to the covariance condition $\text{Cov}(f(X),g(X))=0$. The covariance is computed as:
\begin{align}
	\text{Cov}(f(X),g(X))=\mathbb{E}_{X} A(X-x^0)(X-x^0)^T B^T =A\Sigma B^T
\end{align}
Note that 
\begin{align}
\begin{split}
A\Sigma B^T=0 &\iff \langle v, A\Sigma B^T w \rangle=0 \,\,\forall \,\,v,w \iff \langle A^T v, B^T w \rangle_\Sigma=0 \,\,\forall \,\,v,w \\
&\iff \text{im}(A^T) \perp_\Sigma \text{im}(B^T) \iff \ker(A)^\perp\perp_\Sigma\ker(B)^\perp
\end{split}
\end{align}
\end{proof}

\section{Algorithm}

Algorithm~\ref{alg:rebias} shows the detailed algorithm for solving the minimax problem in equation 3 of the main paper.

\begin{algorithm}
\caption{\methodname training}
\label{alg:rebias}
\begin{algorithmic}
\REPEAT
\FOR{each mini-batch samples $x, y$}
\STATE update $f$ by solving $\underset{f\in F}{\argmin}\,\mathcal{L}(f, x, y) + \lambda\,\text{HSIC}_1(f(x), g(x))$.
\vspace{0.3em}
\STATE update $g$ by solving $\underset{g\in G}{\argmin}\,\mathcal{L}(g, x, y) - \lambda_g\,\text{HSIC}_1(f(x) ,g(x))$.
\ENDFOR
\UNTIL{converge.}
\end{algorithmic}
\end{algorithm}

$\mathcal{L}(f, x, y)$ denotes the original loss for the main task, \eg, the cross entropy loss. We solve the minimax problem in a mini-batch by employing the unbiased finite-sample estimator of HSIC, $\text{HSIC}^{k,l}_1(U,V)$~\citep{unbiasedHSIC}, to measure the independence of $f(X)$ and $g(X)$ in the mini-batch, defined as
\begin{equation}
    \textbf{Unbiased HSIC estimator:}\quad\: \text{HSIC}^{k,l}_1(U,V) =\frac{1}{m(m-3)}\left[\text{tr}(\widetilde{U}\widetilde{V}^T)     + \frac{\mathbf{1}^T \widetilde{U} \mathbf{1} \mathbf{1}^T \widetilde{V}^T \mathbf{1}}{(m-1)(m-2)} - \frac{2}{m-2} \mathbf{1}^T \widetilde{U} \widetilde{V}^T  \mathbf{1} \right]
    \label{eq:unbiased_hsic_supp}
\end{equation}

\section{Implementation Details}
\label{appendix:implementation-details}

\paragraph{Training setup.}
We solve the minimax problem in algorithm~\ref{alg:rebias} through alternating stochastic gradient descents with the ADAM optimiser~\cite{kingma2014adam}. The regularisation parameters $\lambda$ and $\lambda_g$ are set to 1.0 in all experiments. We used the batch sizes (256, 128, 128) for (Biased MNIST, ImageNet, action recognition) experiments. For Biased MNIST, the learning rate is initially set to 0.001 and is decayed by factor 0.1 every 20 epochs. For ImageNet and action recognition, learning rates are initially set to 0.001 and 0.1, respectively, and are decayed by cosine annealing. For action recognition, we use $f(x)$ and $g(x)$ as the output logits for the sake of stable training. For each dataset, we train every method (vanilla, biased, comparison methods, and \methodname) for the same number of epochs. We train the models with (80, 120, 120) epochs for (Biased MNIST, ImageNet, action recognition) experiments. All experiments are implemented using PyTorch~\citep{paszke2019pytorch}.

In addition, we observe that a better optimiser than ADAM, \eg AdamP~\cite{heo2020adamp}, helps performance improvements in Biased MNIST and ImageNet experiments. Table~\ref{table:adamp} shows the results from the reference paper \cite{heo2020adamp}. For future researches, we recommend using AdamP for better optimisation stability.

\definecolor{darkergreen}{RGB}{21, 152, 56}
\newcommand\greenp[1]{\textcolor{darkergreen}{(#1)}}
\definecolor{red2}{RGB}{252, 54, 65}
\newcommand\redp[1]{\textcolor{red2}{(#1)}}
\newcommand\greenpscript[1]{\scriptsize\greenp{#1}}
\newcommand\redpscript[1]{\scriptsize\redp{#1}}

\begin{table}[h]
\setlength{\tabcolsep}{3.5pt}
\centering
\caption{\small \textbf{AdamP experiments.} Biased MNIST and 9-Class ImageNet benchmarks with AdamP~\cite{heo2020adamp}.}
\label{table:adamp}
\vspace{.5em}
\small
\begin{tabular}{@{}lccccccccc@{}}
\toprule
\multirow{2}{*}{Optimizer} & \multicolumn{5}{c}{Biased MNIST Unbiased acc. at $\rho$} & & \multicolumn{3}{c}{9-Class ImageNet} \\ \cmidrule(l){2-6} \cmidrule(l){8-10}
      & .999 & .997 & .995 & .990 & avg. & & Biased     & UnBiased     & IN-A     \\ \midrule
Adam~\cite{kingma2014adam}  & 22.9           & 63.0           & 74.9           & 87.0       & 61.9          & & 93.8      & 92.6        & 31.2    \\
AdamP~\cite{heo2020adamp} & \textbf{30.5 \greenpscript{+7.5}}           & \textbf{70.9 \greenpscript{+7.9}}           & \textbf{80.9 \greenpscript{+6.0}}           & \textbf{89.6 \greenpscript{+2.6}}  & \textbf{68.0 \greenpscript{+6.0}}    &      & \textbf{95.2 \greenpscript{+1.4}}      & \textbf{94.5 \greenpscript{+1.8}}        & \textbf{32.9 \greenpscript{+1.7}}    \\ \bottomrule
\end{tabular}
\end{table}

\paragraph{Architecture details for action recognition.}
The \arch{3D-ResNet}~\cite{tran2019video, feichtenhofer2019slowfast} architecture is widely used in various video understanding tasks. We choose \arch{3D-ResNet18} and \arch{2D-ResNet18} as F and G of our method, respectively. The architectural details are in Table~\ref{table:resnet3d_arch_supp}.

\paragraph{Training details for comparison methods.}
For training \method{LearnedMixin}, we pre-train and fix $G$ before training $F$, as done in the original paper. We pre-train $G$ for 5 epochs for Biased MNIST, and 30 epochs for ImageNet and action recognition. For training \method{RUBi}, we update $F$ and $G$ simultaneously without pre-training $G$, as done in the original paper. For training \method{HEX}, we substitute $G$ with the neural grey-level co-occurrence matrix (NGLCM) to represent the ``superficial statistics''. For \method{StylisedImageNet}, we augment 9-class ImageNet with its stylised version (\ie, twice the original dataset size), while maintaining the training setup as identical.

\newcommand{\x}{{$\times$}}
\newcommand{\blockf}[2]{\multirow{2}{*}{\(\left[\begin{array}{c}\text{\textbf{3}$\times$3$\times$3, #1}\\[-.1em] \text{1$\times$3$\times$3, #1}\\[-.1em]\end{array}\right]\)$\times$#2}}
\newcommand{\blockg}[2]{\multirow{2}{*}{\(\left[\begin{array}{c}\text{\textbf{1}$\times$3$\times$3, #1}\\[-.1em] \text{1$\times$3$\times$3, #1}\\[-.1em]\end{array}\right]\)$\times$#2}}
\newcommand{\outputsize}[1]{\multirow{2}{*}{#1}}

\begin{table}[t]
\small
\centering
\begin{tabular}{@{}c|c|c|c@{}}
\toprule
layer name & \arch{3D-ResNet18} & \arch{2D-ResNet18}        & output size (T\x C\x H\x W)          \\ \midrule
input      & input video                                    &  input video            & 8\x3\x224\x224                    \\ \midrule
conv1      & \begin{tabular}[c]{@{}c@{}}\textbf{3}\x7\x7, 32\\ stride=2  \end{tabular} & \begin{tabular}[c]{@{}c@{}}\textbf{1}\x7\x7, 32\\ stride=2  \end{tabular} & 8\x32\x112\x112           \\ \midrule
pool1      & \begin{tabular}[c]{@{}c@{}}1\x3\x7, max\\ stride=2  \end{tabular} & \begin{tabular}[c]{@{}c@{}}1\x3\x7, max\\ stride=2  \end{tabular} & 8\x32\x56\x56             \\ \midrule
\multirow{2}{*}{resblock2}  &       \blockf{{32}}{2}      & \blockg{{32}}{2} &   \outputsize{8\x32\x56\x56}                         \\ 
			&	&  & \\\midrule
\multirow{2}{*}{resblock3}  &       \blockf{{64}}{2}    &  \blockg{{32}}{2}  &    \outputsize{8\x64\x28\x28}                         \\ 
			&	&  & \\\midrule
\multirow{2}{*}{resblock4}  &       \blockf{{128}}{2}    &  \blockg{{32}}{2}  &   \outputsize{8\x128\x14\x14}                          \\ 
			&	&  & \\\midrule
\multirow{2}{*}{resblock5}  &       \blockf{{256}}{2}     &  \blockg{{32}}{2}  &   \outputsize{8\x256\x7\x7}                          \\ 
			&	&  & \\\midrule				
final           & \begin{tabular}[c]{@{}c@{}}global avg. pool\\ fc \end{tabular} & \begin{tabular}[c]{@{}c@{}}global avg. pool\\ fc \end{tabular}  &       \# classes                \\ \bottomrule
\end{tabular}
   \caption{\textbf{Network architecture for action recognisers.} The kernel dimensions of convolution layers are denoted as $\left[\text{T}\times\text{W}\times\text{H}, \text{C}\right]$ along temporal (T), width (W), height (H), and channel (C) axes, respectively. The main difference between \arch{3D-ResNet18} and \arch{2D-ResNet18} is the temporal kernel size.}
    \label{table:resnet3d_arch_supp}
\end{table}

\section{Standard Errors in Experimental Results}

We report the standard errors of the main paper results in Table~\ref{table:mnist-experiment-supp} (Biased MNIST), Table~\ref{table:imagenet-experiment-supp} (ImageNet) and Table~\ref{table:action-experiment-supp} (action recognition).

\begin{table*}[ht!]
\centering
\setlength\tabcolsep{0.4em}
\small
\begin{tabular}{cccccccg}
$\rho$ & & { Vanilla} & { Biased} & {\method{HEX}} & {\method{LearnedMixin}} & {\method{RUBi}} & {\methodname (ours)} \\
\cline{1-1} \cline{3-8}
& \vspace{-1em} \\
\cline{1-1} \cline{3-8}
& \vspace{-1em} \\
.999             && 10.4 $\pm$ 0.5 & 10. $\pm$ 0. &      10.8 $\pm$ 0.4 & 12.1 $\pm$ 0.8 & 13.7 $\pm$ 0.7 & \textbf{22.7} $\pm$ 0.4 \\
.997             && 33.4 $\pm$ 12.1 & 10. $\pm$ 0. & 16.6 $\pm$ 0.8 & 50.2 $\pm$ 4.5 & 43.0 $\pm$ 1.1 & \textbf{64.2} $\pm$ 0.8 \\
.995             && 72.1 $\pm$ 1.9 & 10. $\pm$ 0. & 19.7 $\pm$ 1.9 & 78.2 $\pm$ 0.7 & \textbf{90.4} $\pm$ 0.4 & 76.0 $\pm$ 0.6 \\
.990          && 89.1 $\pm$ 0.1 & 10. $\pm$ 0.  & 24.7  $\pm$ 1.6 & 88.3 $\pm$ 0.7 & \textbf{93.6} $\pm$ 0.4 & 88.1 $\pm$ 0.6 \\
\cline{1-1} \cline{3-8}
& \vspace{-1em} \\
avg.           && 51.2            & 10.    & 18.0          & 57.2       & 60.2 & \textbf{62.7} \\
\cline{1-1} \cline{3-8}
& \vspace{-1em} \\
\end{tabular}
\caption{\small\textbf{Unbiased accuracy on Biased MNIST.} Unbiased accuracies on varying train correlation $\rho$. Means and standard errors are computed over three independent runs.}
\label{table:mnist-experiment-supp}
\end{table*}

\begin{table}[ht!]
\centering
\setlength\tabcolsep{0.4em}
\small
\begin{tabular}{lcccc}
Model description && Biased  & Unbiased & IN-A \\ \cline{1-1} \cline{3-5}
\vspace{-1em} & \\ 
\cline{1-1} \cline{3-5}
\vspace{-1em} & \\
Vanilla {(\arch{ResNet18})} && 90.8 $\pm$ 0.6 &  88.8 $\pm$ 0.6 & 24.9 $\pm$ 1.1 \\  
Biased {(\arch{BagNet18})} && 67.7 $\pm$ 0.3 &  65.9 $\pm$ 0.3 & 18.8 $\pm$ 1.1 \\   
\cline{1-1} \cline{3-5}
\vspace{-1em} & \\
{\method{StylisedIN}} {\cite{StylisedImageNet}} && 88.4 $\pm$ 0.5 & 86.6 $\pm$ 0.6 & 24.6 $\pm$ 1.4\\
{\method{LearnedMixin}} {\cite{clark2019don}} && 64.1 $\pm$ 4.0 & 62.7 $\pm$ 3.1 & 15.0 $\pm$ 1.6 \\ 
{\method{RUBi}} {\cite{cadene2019rubi}} && 90.5$ \pm$ 0.3 & 88.6 $\pm$ 0.4 & 27.7 $\pm$ 2.1 \\
\cline{1-1} \cline{3-5}
\vspace{-1em} & \\
\cellcolor{Gray}\methodname (ours) && \cellcolor{Gray}\textbf{91.9} $\pm$ 1.7 & \cellcolor{Gray}\textbf{90.5} $\pm$ 1.7 & \cellcolor{Gray}\textbf{29.6} $\pm$ 1.6 \\
\cline{1-1} \cline{3-5}
\end{tabular}
\caption{\small\textbf{ImageNet results.} We show results corresponding to $F=\arch{ResNet18}$ and $G=\arch{BagNet18}$. IN-A indicates ImageNet-A. Means and standard errors are computed over three independent runs.}
\label{table:imagenet-experiment-supp}
\end{table}

\begin{table}[ht!]
\centering
\setlength\tabcolsep{0.4em}
\small
\begin{tabular}{lcccc}
&& Biased     & Unbiased   \\
Model description && (Kinetics) & (Mimetics) \\
\cline{1-1} \cline{3-4}
\vspace{-1em} & \\ 
\cline{1-1} \cline{3-4}
\vspace{-1em} & \\
Vanilla {(\arch{3D-ResNet18})} &&  54.5 $\pm$ 3.2 &  18.9 $\pm$ 0.4\\  
Biased {(\arch{2D-ResNet18})} &&  50.7 $\pm$ 3.3 &  18.4 $\pm$ 2.3 \\  
\cline{1-1} \cline{3-4}
& \vspace{-1em} \\
{\method{LearnedMixin}} {\cite{clark2019don}} && 12.3 $\pm$ 2.3 & 11.4 $\pm$ 0.4 \\
{\method{RUBi}} {\cite{cadene2019rubi}}  && 22.4 $\pm$ 2.0 & 13.4 $\pm$ 1.5 \\
\cline{1-1} \cline{3-4}
& \vspace{-1em} \\
\cellcolor{Gray} \methodname (ours) && \cellcolor{Gray}\textbf{55.8} $\pm$ 3.1 & \cellcolor{Gray}\textbf{22.4} $\pm$ 1.3 \\ 
\cline{1-1} \cline{3-4}
\end{tabular}
\caption{\textbf{Action recognition results.} We show results corresponding to $F=\arch{3D-ResNet18}$ and $G=\arch{2D-ResNet18}$ with baseline comparisons. Top-1 accuracies are reported. Means and standard errors are computed over three independent runs.}
\label{table:action-experiment-supp}
\end{table}

\section{Decision Boundary Visualisation for Toy Experiment}
\label{appendix:decision-boundary}

\begin{figure}[ht!]
	\centering
	\begin{tabular}{ccc}
	    \includegraphics[width=0.2\columnwidth]{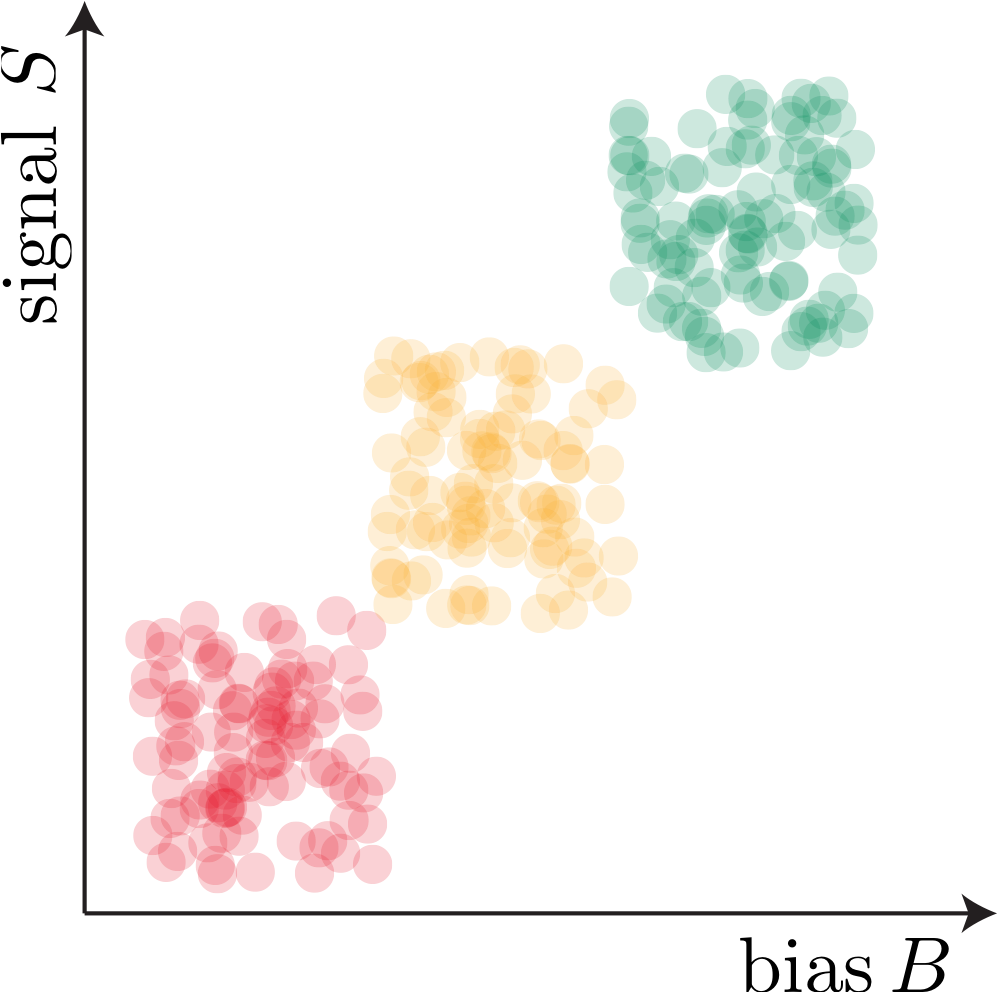} &  
	    \includegraphics[width=0.2\columnwidth]{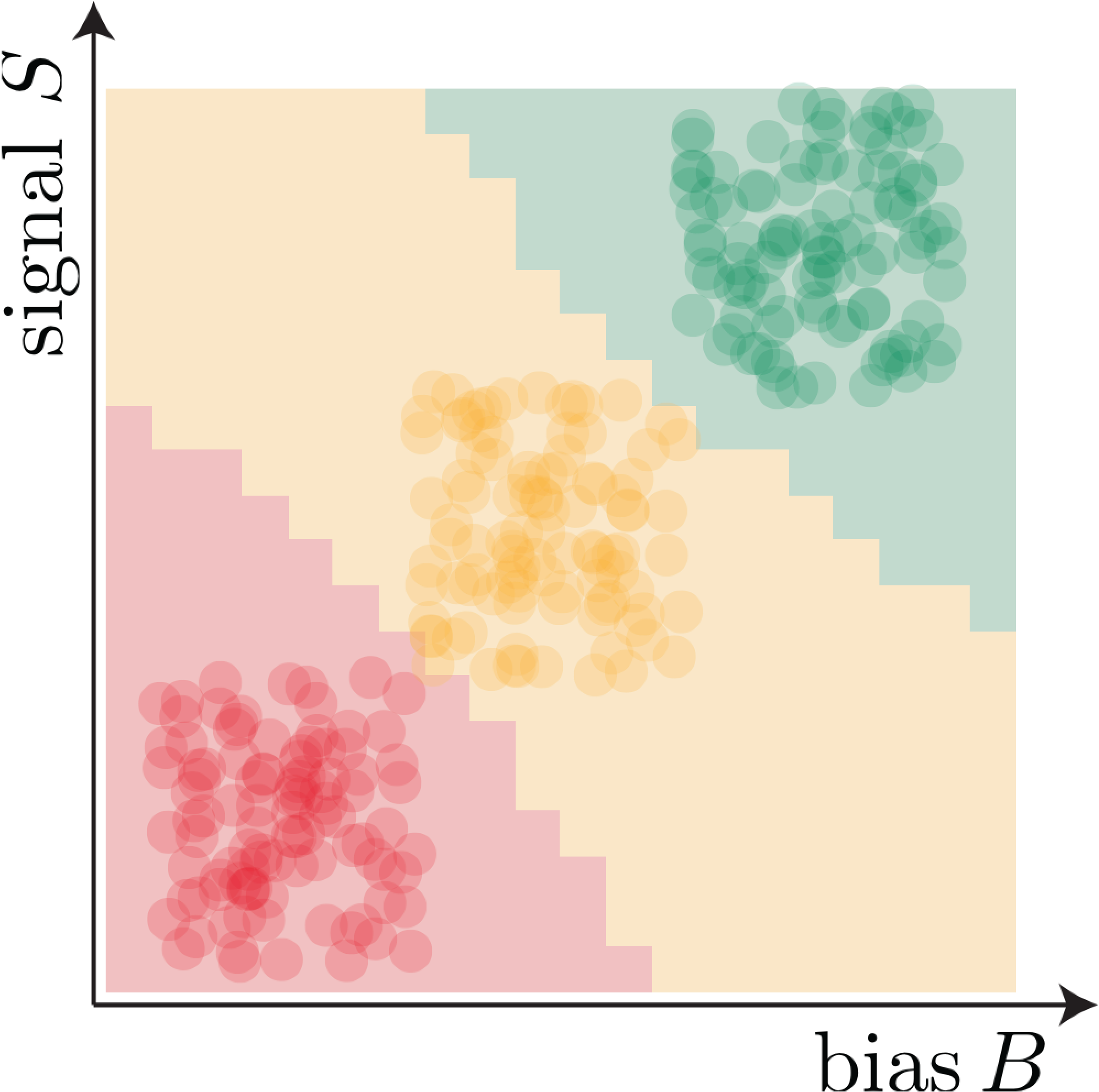} &
	    \includegraphics[width=0.2\columnwidth]{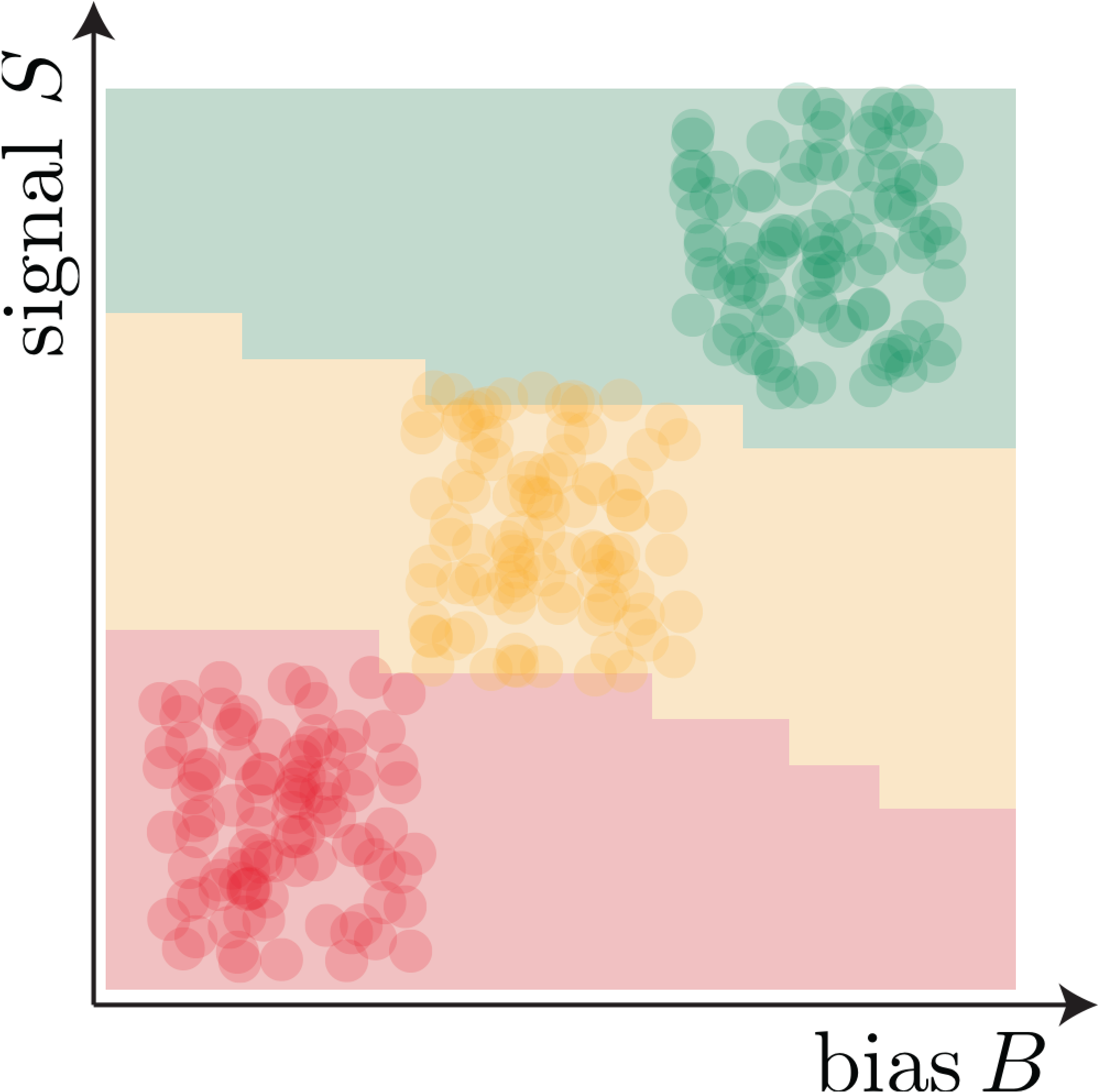} \\
	    Training data & Vanilla (baseline) & \methodname (ours)
	\end{tabular}
	\caption{\small \textbf{Decision boundaries of toy data}. GIF animations are at \href{https://anonymous.4open.science/r/49609e91-5a76-48fd-86c8-385b45b34dee/}{\color{blue}{anonymous link}}.}
	\label{fig:toy-decision-boundary}
\end{figure}

We show the decision boundaries of the toy experiment (\S3.2) in Figure~\ref{fig:toy-decision-boundary}.

\clearpage
\section{Texture Clustering on ImageNet}
\label{appendix:clustering}

In our ImageNet experiments (\S4.3), we have obtained the proxy ground truths for texture bias using texture feature clustering. We extract the texture features from images by computing the gram matrices of low-layer feature maps, as done in texturisation methods~\citep{gatys2015texture, johnson2016perceptual}, to capture the edge and colour cues. Specifically, we use the feature maps from layer \arch{relu1\_2} of the ImageNet pre-trained \arch{VGG16}~\citep{simonyan2014very}. 

To approximate the texture ground truth labels, we cluster the texture features of 9-Class ImageNet data. We use the mini-batch $k$-means algorithm with $k=9$ and batch size 1024. As k-means clustering is non-convex, we repeat the ImageNet experiments with three different texture clustering results, each with different initialisation. We report the averaged performances across the three trials.

\begin{figure}[ht!]
    \centering
    \includegraphics[width=0.65\columnwidth]{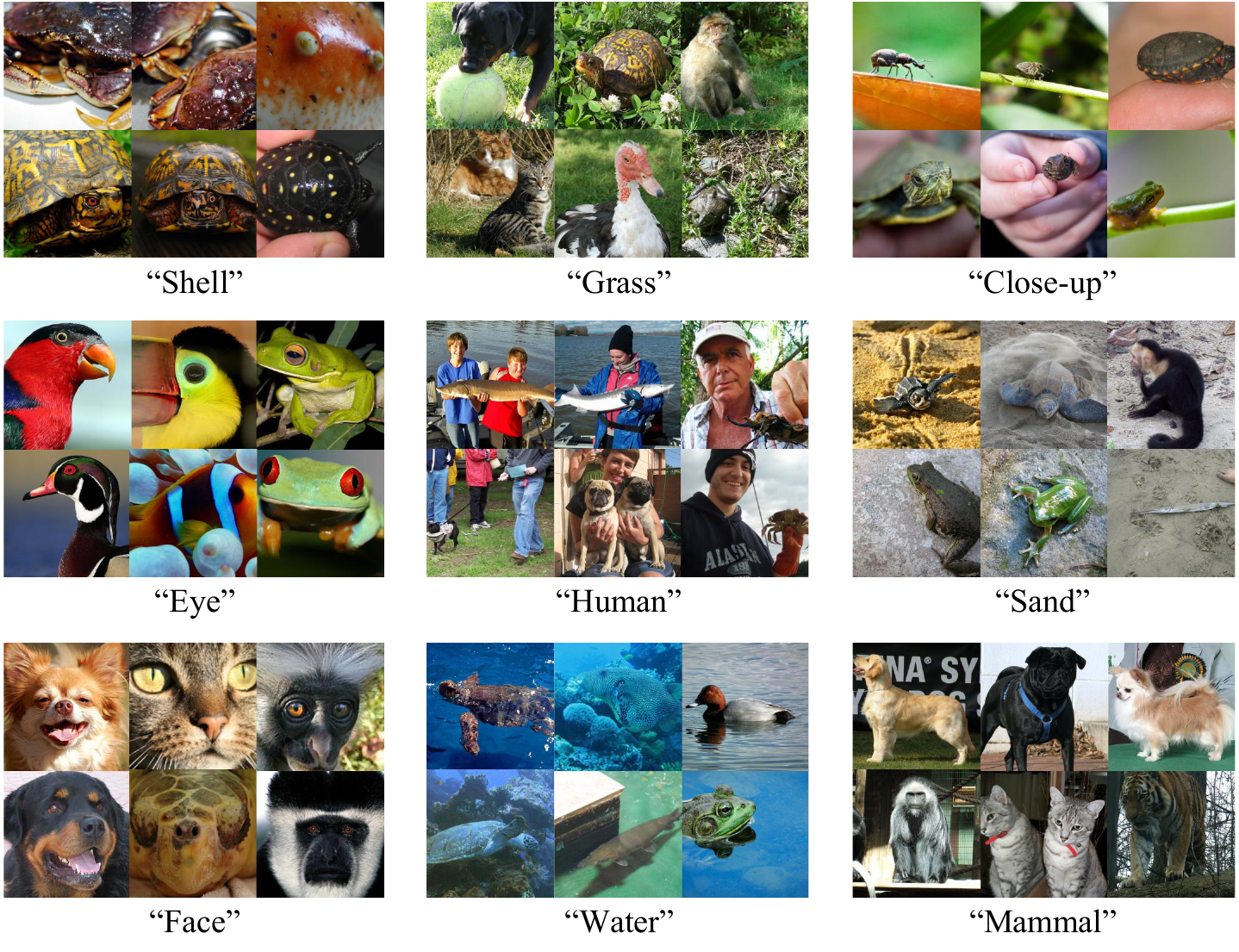}
    \vspace{-1.em}
    \caption{\small\textbf{Texture clusters.} Example images from texture clusters. Each cluster is named according to the common texture pattern.}
    \label{fig:all_cluster}
\end{figure}

\begin{figure}[ht!]
    \centering
    \includegraphics[width=.9\columnwidth]{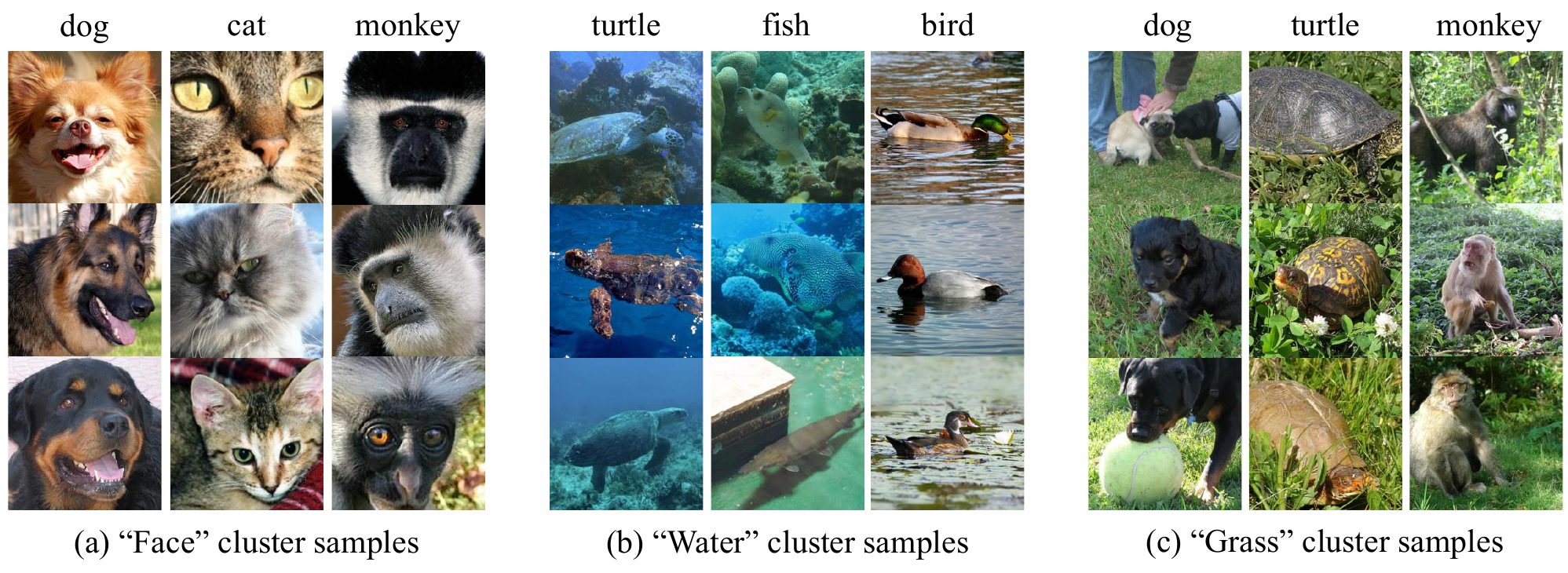}
    \vspace{-1.em}
    \caption{\small\textbf{Dominant texture-class correlations.} For each cluster, we visualise its top-3 correlated classes.}
    \label{fig:texture_clusters}
\end{figure}

We show an example texture clustering in Figure~\ref{fig:all_cluster}. Clusters capture similar texture patterns. The texture clusters exhibit strong correlations with semantic classes. See Figure~\ref{fig:texture_clusters} for the top-3 correlated classes per cluster. For example, the ``water'' texture is strongly associated with the turtle, fish, and bird classes. In the presence of such bias-class correlations, the model is motivated to take the bias shortcut by utilising the texture cue for recognition. In this case, the model shows sub-optimal performances on unusual class-texture combinations (\eg, crab on grass texture).

\section{Further Analysis on Texture Bias of ImageNet-Trained Models}
\label{appendix:bias-data-models}

\begin{figure}[h]
\centering
\begin{subfigure}[b]{0.497\textwidth}
\centering
    \includegraphics[width=1\linewidth]{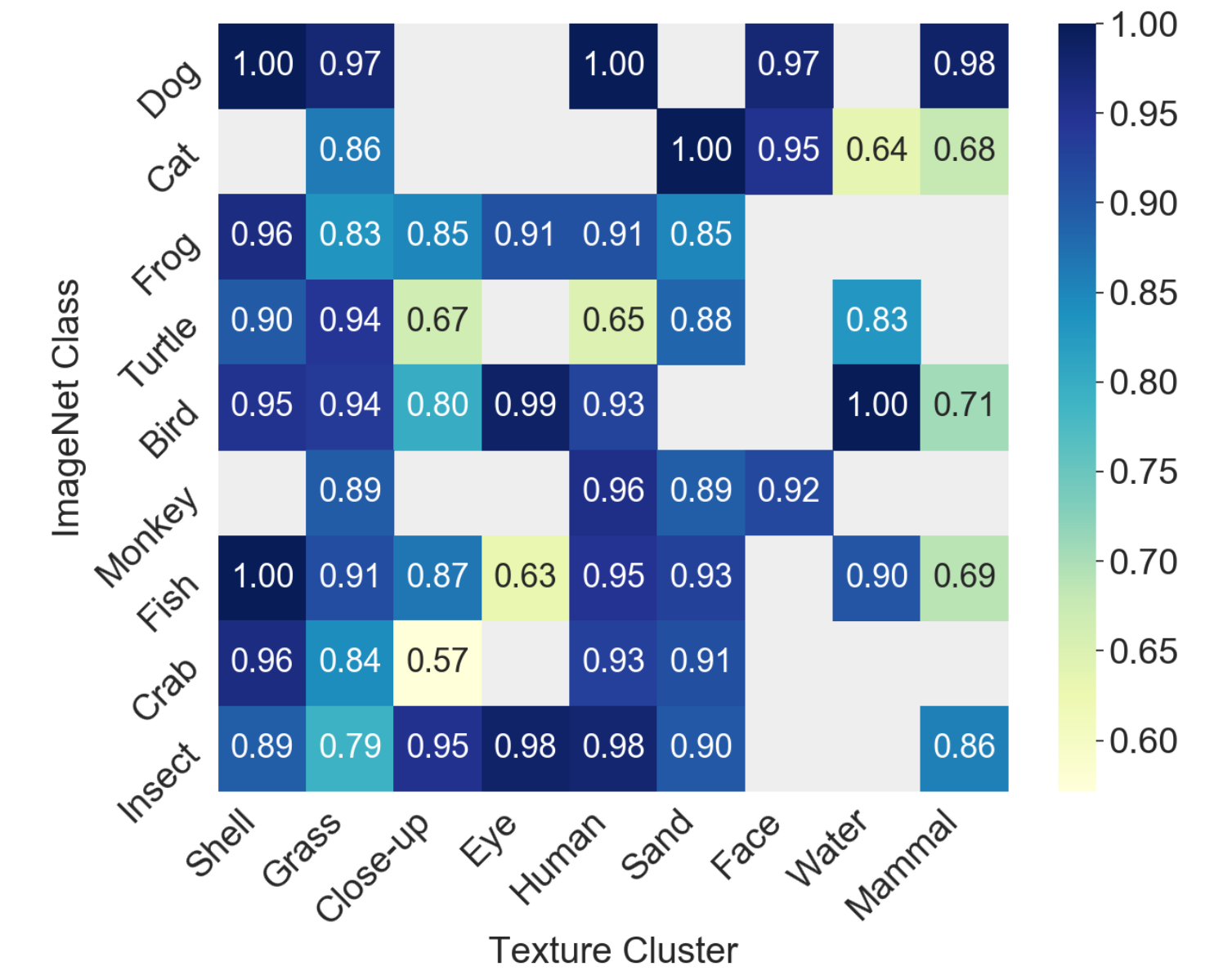}
    \caption{Vanilla trained \arch{ResNet18}.}
    \label{fig:vanilla_cm_imagenet}
\end{subfigure}
\begin{subfigure}[b]{0.497\textwidth}
\centering
    \includegraphics[width=1\linewidth]{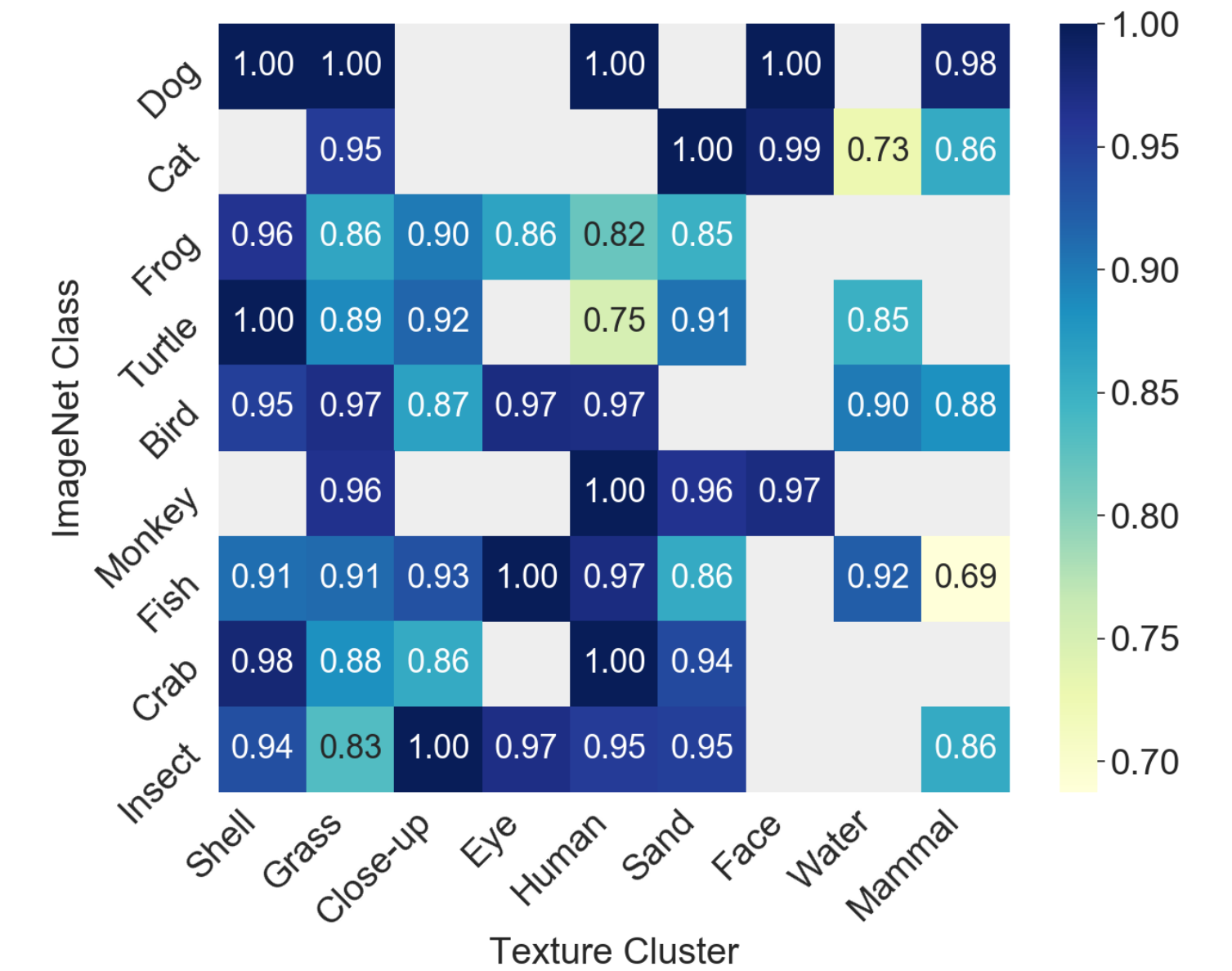}
    \caption{\methodname trained \arch{ResNet18}.}
    \label{fig:ours_cm_imagenet}
\end{subfigure}
    \vspace{-2em}
    \caption{\small\textbf{Texture-class-wise accuracies on ImageNet.}
    For every texture-class pair $(B,Y)=(b,y)$, corresponding accuracy is visualised. We ignore cells with population less than 10 (masked in gray).}
    \label{fig:cluster_class_correlation_analysis}
\end{figure}

We further analyse the texture bias of the models trained on ImageNet (\S4.3). Figure~\ref{fig:cluster_class_correlation_analysis} shows the texture-class-wise accuracies of the vanilla-trained \arch{ResNet18} and \methodname-trained \arch{ResNet18}. To quantitatively measure the texture biases, we count the number of samples for each texture-class pair $(c,y)$ (``population'' $\text{Pop}(c,y)$ in the main paper). For each class, we define the \textit{dominant texture cluster} as the largest cluster that contains $>30\%$ of the class samples. 4 out of 9 classes have the dominant texture cluster: (``Dog'', ``Face''), (``Cat'', ``Face''), (``Bird'', ``Eye''), and (``Monkey'', ``Face''). 

We measure the average accuracy over classes with \textit{dominant texture clusters} (biased classes) and the average over the other classes. We observe that \arch{ResNet18} shows higher accuracy on biased classes ($90.6\%$) than on less biased classes ($86.3\%$), signifying its bias towards texture. On the other hand, \methodname achieves similar accuracies $94.8\%$ (biased classes) and $90.4\%$ (unbiased classes). We stress that \methodname overcomes the bias and enhances generalisation across distributions even if the training dataset itself is biased.

\begin{figure}[ht!]
\centering
\small
\begin{tabular}{ccccc}
	Kinetics & Mimetics && Kinetics & Mimetics \\
	\includegraphics[width=0.2\columnwidth]{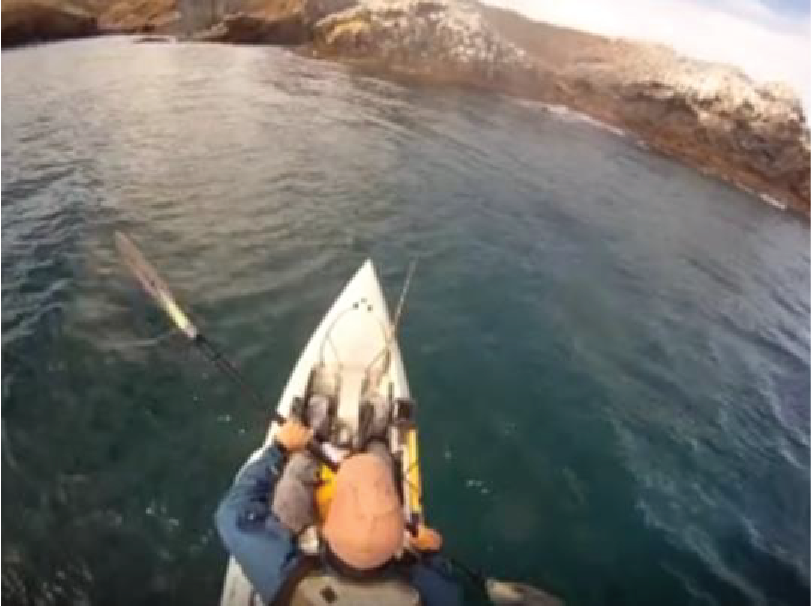} &  
	\includegraphics[width=0.2\columnwidth]{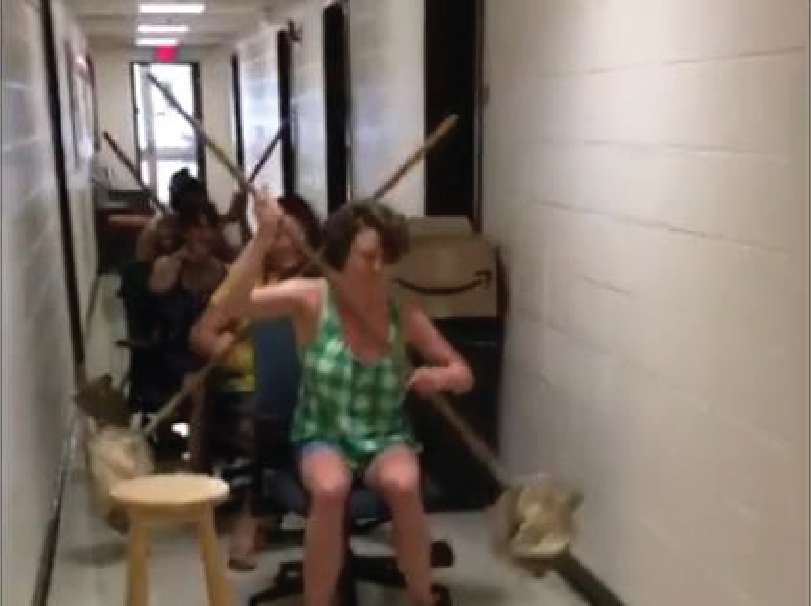} &&
	\includegraphics[width=0.2\columnwidth]{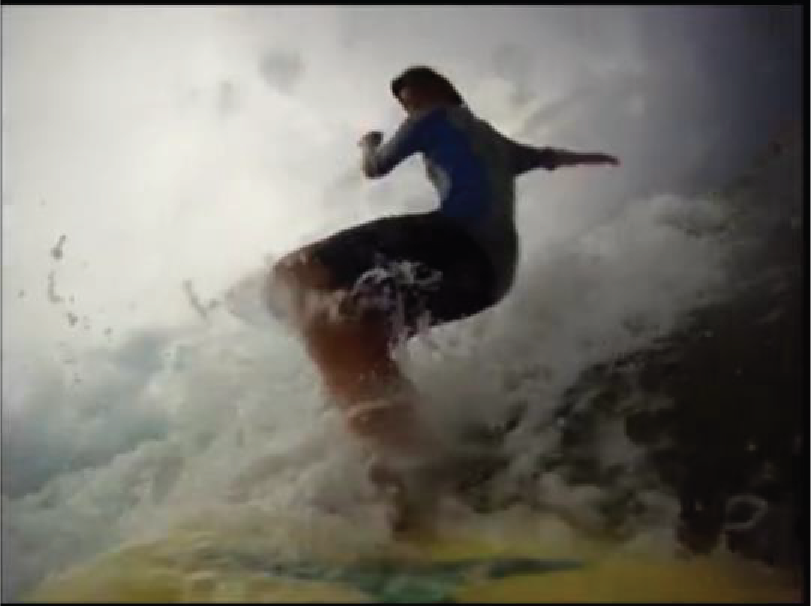} &  
	\includegraphics[width=0.2\columnwidth]{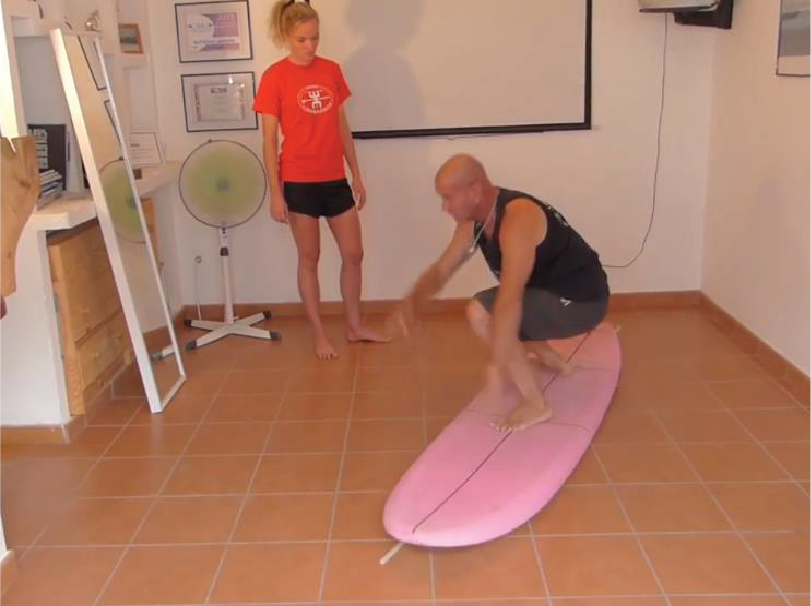} \\
	\multicolumn{2}{c}{Canoeing or kayaking} && \multicolumn{2}{c}{Surfing water}
\end{tabular}
   \caption{\textbf{Kinetics and Mimetics datasets.} We show examples of two classes, ``canoeing or kayaking'' and ``surfing water''. Kinetics samples are biased towards certain scene contexts (\eg ocean), but Mimetics is relatively free from such context biases.}
\label{fig:kinetics_mimetics}
\end{figure}

\vspace{-1em}
\section{Action Recognition Datasets}
\label{appendix:action_recogition_datasets}

We provide an overview of the action recognition datasets.
Kinetics dataset~\cite{carreira2017quo} has 300K videos with 400 action classes. Mimetics dataset~\cite{weinzaepfel2019mimetics} has 713 videos with 50 action classes (subset of the Kinetics classes). To simplify our investigation, we have sub-sampled 10 common classes from Kinetics and Mimetics: \textit{``canoeing or kayaking'', ``climbing a rope'', ``driving car'', ``golf driving'', ``opening bottle'', ``playing piano'', ``playing volleyball'', ``shooting goal (soccer)'', ``surfing water'', and ``writing''}. Examples of Kinetics and Mimetics are shown in Figure~\ref{fig:kinetics_mimetics}. While Kinetics samples are biased towards static cues like scene and objects, Mimetics are relatively free of such correlations. Mimetics is thus a suitable benchmark for validating the cross-bias generalisation performances.

%% file: main.bbl
\begin{thebibliography}{51}
\providecommand{\natexlab}[1]{#1}
\providecommand{\url}[1]{\texttt{#1}}
\expandafter\ifx\csname urlstyle\endcsname\relax
  \providecommand{\doi}[1]{doi: #1}\else
  \providecommand{\doi}{doi: \begingroup \urlstyle{rm}\Url}\fi

\bibitem[Agarwal et~al.(2019)Agarwal, Shetty, and Fritz]{agarwal2019towards}
Agarwal, V., Shetty, R., and Fritz, M.
\newblock Towards causal vqa: Revealing and reducing spurious correlations by
  invariant and covariant semantic editing.
\newblock \emph{arXiv preprint arXiv:1912.07538}, 2019.

\bibitem[Agrawal et~al.(2018)Agrawal, Batra, Parikh, and
  Kembhavi]{agrawal2018don}
Agrawal, A., Batra, D., Parikh, D., and Kembhavi, A.
\newblock Don't just assume; look and answer: Overcoming priors for visual
  question answering.
\newblock In \emph{Proceedings of the IEEE Conference on Computer Vision and
  Pattern Recognition}, pp.\  4971--4980, 2018.

\bibitem[Ben-David et~al.(2007)Ben-David, Blitzer, Crammer, and
  Pereira]{ben2007analysis}
Ben-David, S., Blitzer, J., Crammer, K., and Pereira, F.
\newblock Analysis of representations for domain adaptation.
\newblock In \emph{Advances in neural information processing systems}, pp.\
  137--144, 2007.

\bibitem[Brendel \& Bethge(2019)Brendel and Bethge]{BagNet}
Brendel, W. and Bethge, M.
\newblock Approximating {CNN}s with bag-of-local-features models works
  surprisingly well on imagenet.
\newblock In \emph{International Conference on Learning Representations}, 2019.
\newblock URL \url{https://openreview.net/forum?id=SkfMWhAqYQ}.

\bibitem[Cadene et~al.(2019)Cadene, Dancette, Cord, Parikh,
  et~al.]{cadene2019rubi}
Cadene, R., Dancette, C., Cord, M., Parikh, D., et~al.
\newblock Rubi: Reducing unimodal biases for visual question answering.
\newblock In \emph{Advances in Neural Information Processing Systems}, pp.\
  839--850, 2019.

\bibitem[Carreira \& Zisserman(2017)Carreira and Zisserman]{carreira2017quo}
Carreira, J. and Zisserman, A.
\newblock Quo vadis, action recognition? a new model and the kinetics dataset.
\newblock In \emph{proceedings of the IEEE Conference on Computer Vision and
  Pattern Recognition}, pp.\  6299--6308, 2017.

\bibitem[Choi et~al.(2019)Choi, Gao, Messou, and Huang]{choi2019can}
Choi, J., Gao, C., Messou, J.~C., and Huang, J.-B.
\newblock Why can't i dance in the mall? learning to mitigate scene bias in
  action recognition.
\newblock In \emph{Advances in Neural Information Processing Systems}, pp.\
  853--865, 2019.

\bibitem[Clark et~al.(2019)Clark, Yatskar, and Zettlemoyer]{clark2019don}
Clark, C., Yatskar, M., and Zettlemoyer, L.
\newblock Don't take the easy way out: Ensemble based methods for avoiding
  known dataset biases.
\newblock In \emph{Proceedings of the 2019 Conference on Empirical Methods in
  Natural Language Processing and the 9th International Joint Conference on
  Natural Language Processing (EMNLP-IJCNLP)}, pp.\  4069--4082, 2019.

\bibitem[Creager et~al.(2019)Creager, Madras, Jacobsen, Weis, Swersky, Pitassi,
  and Zemel]{creager2019flexibly}
Creager, E., Madras, D., Jacobsen, J.-H., Weis, M., Swersky, K., Pitassi, T.,
  and Zemel, R.
\newblock Flexibly fair representation learning by disentanglement.
\newblock In \emph{International Conference on Machine Learning}, pp.\
  1436--1445, 2019.

\bibitem[Feichtenhofer et~al.(2019)Feichtenhofer, Fan, Malik, and
  He]{feichtenhofer2019slowfast}
Feichtenhofer, C., Fan, H., Malik, J., and He, K.
\newblock Slowfast networks for video recognition.
\newblock In \emph{Proceedings of the IEEE International Conference on Computer
  Vision}, pp.\  6202--6211, 2019.

\bibitem[Gatys et~al.(2015)Gatys, Ecker, and Bethge]{gatys2015texture}
Gatys, L., Ecker, A.~S., and Bethge, M.
\newblock Texture synthesis using convolutional neural networks.
\newblock In \emph{Advances in neural information processing systems}, pp.\
  262--270, 2015.

\bibitem[Geirhos et~al.(2019)Geirhos, Rubisch, Michaelis, Bethge, Wichmann, and
  Brendel]{StylisedImageNet}
Geirhos, R., Rubisch, P., Michaelis, C., Bethge, M., Wichmann, F.~A., and
  Brendel, W.
\newblock Imagenet-trained {CNN}s are biased towards texture; increasing shape
  bias improves accuracy and robustness.
\newblock In \emph{International Conference on Learning Representations}, 2019.
\newblock URL \url{https://openreview.net/forum?id=Bygh9j09KX}.

\bibitem[Gretton et~al.(2005)Gretton, Bousquet, Smola, and Sch{\"o}lkopf]{HSIC}
Gretton, A., Bousquet, O., Smola, A., and Sch{\"o}lkopf, B.
\newblock Measuring statistical dependence with hilbert-schmidt norms.
\newblock In \emph{International conference on algorithmic learning theory},
  pp.\  63--77. Springer, 2005.

\bibitem[Gretton et~al.(2008)Gretton, Fukumizu, Teo, Song, Sch{\"o}lkopf, and
  Smola]{HSICTest}
Gretton, A., Fukumizu, K., Teo, C.~H., Song, L., Sch{\"o}lkopf, B., and Smola,
  A.~J.
\newblock A kernel statistical test of independence.
\newblock In \emph{Advances in neural information processing systems}, pp.\
  585--592, 2008.

\bibitem[Gururangan et~al.(2018)Gururangan, Swayamdipta, Levy, Schwartz,
  Bowman, and Smith]{gururangan2018naacl}
Gururangan, S., Swayamdipta, S., Levy, O., Schwartz, R., Bowman, S., and Smith,
  N.~A.
\newblock Annotation artifacts in natural language inference data.
\newblock In \emph{Proceedings of the 2018 Conference of the North {A}merican
  Chapter of the Association for Computational Linguistics}, pp.\  107--112,
  New Orleans, Louisiana, June 2018. Association for Computational Linguistics.
\newblock URL \url{https://www.aclweb.org/anthology/N18-2017}.

\bibitem[Hardt et~al.(2016)Hardt, Price, Srebro, et~al.]{Separation}
Hardt, M., Price, E., Srebro, N., et~al.
\newblock Equality of opportunity in supervised learning.
\newblock In \emph{Advances in neural information processing systems}, pp.\
  3315--3323, 2016.

\bibitem[He et~al.(2016)He, Zhang, Ren, and Sun]{ResNet}
He, K., Zhang, X., Ren, S., and Sun, J.
\newblock Deep residual learning for image recognition.
\newblock In \emph{Proceedings of the IEEE conference on computer vision and
  pattern recognition}, pp.\  770--778, 2016.

\bibitem[Hendrycks \& Dietterich(2019)Hendrycks and
  Dietterich]{hendrycks2018imagenet-c}
Hendrycks, D. and Dietterich, T.
\newblock Benchmarking neural network robustness to common corruptions and
  perturbations.
\newblock In \emph{International Conference on Learning Representations}, 2019.
\newblock URL \url{https://openreview.net/forum?id=HJz6tiCqYm}.

\bibitem[Hendrycks et~al.(2019)Hendrycks, Zhao, Basart, Steinhardt, and
  Song]{hendrycks2019natural}
Hendrycks, D., Zhao, K., Basart, S., Steinhardt, J., and Song, D.
\newblock Natural adversarial examples.
\newblock \emph{arXiv preprint arXiv:1907.07174}, 2019.

\bibitem[Heo et~al.(2020)Heo, Chun, Oh, Han, Yun, Uh, and Ha]{heo2020adamp}
Heo, B., Chun, S., Oh, S.~J., Han, D., Yun, S., Uh, Y., and Ha, J.-W.
\newblock Slowing down the weight norm increase in momentum-based optimizers.
\newblock \emph{arXiv preprint arXiv:2006.08217}, 2020.

\bibitem[Ilyas et~al.(2019)Ilyas, Santurkar, Tsipras, Engstrom, Tran, and
  Madry]{ilyas2019adversarial}
Ilyas, A., Santurkar, S., Tsipras, D., Engstrom, L., Tran, B., and Madry, A.
\newblock Adversarial examples are not bugs, they are features.
\newblock In \emph{Advances in Neural Information Processing Systems}, pp.\
  125--136, 2019.

\bibitem[Ioffe \& Szegedy(2015)Ioffe and Szegedy]{batchnorm}
Ioffe, S. and Szegedy, C.
\newblock Batch normalization: Accelerating deep network training by reducing
  internal covariate shift.
\newblock In Bach, F. and Blei, D. (eds.), \emph{Proceedings of the 32nd
  International Conference on Machine Learning}, volume~37 of \emph{Proceedings
  of Machine Learning Research}, pp.\  448--456, Lille, France, 07--09 Jul
  2015. PMLR.

\bibitem[Johnson et~al.(2016)Johnson, Alahi, and
  Fei-Fei]{johnson2016perceptual}
Johnson, J., Alahi, A., and Fei-Fei, L.
\newblock Perceptual losses for real-time style transfer and super-resolution.
\newblock In \emph{European conference on computer vision}, pp.\  694--711.
  Springer, 2016.

\bibitem[Kim et~al.(2019)Kim, Kim, Kim, Kim, and Kim]{kim2019learning}
Kim, B., Kim, H., Kim, K., Kim, S., and Kim, J.
\newblock Learning not to learn: Training deep neural networks with biased
  data.
\newblock In \emph{Proceedings of the IEEE Conference on Computer Vision and
  Pattern Recognition}, pp.\  9012--9020, 2019.

\bibitem[Kim et~al.(2018)Kim, Kim, Seo, Kim, Park, Park, Jo, Kim, Yang, Kim,
  et~al.]{kim2018nsml}
Kim, H., Kim, M., Seo, D., Kim, J., Park, H., Park, S., Jo, H., Kim, K., Yang,
  Y., Kim, Y., et~al.
\newblock Nsml: Meet the mlaas platform with a real-world case study.
\newblock \emph{arXiv preprint arXiv:1810.09957}, 2018.

\bibitem[Kingma \& Ba(2015)Kingma and Ba]{kingma2014adam}
Kingma, D.~P. and Ba, J.
\newblock Adam: A method for stochastic optimization.
\newblock In \emph{International Conference on Learning Representations}, 2015.

\bibitem[Kornblith et~al.(2019)Kornblith, Norouzi, Lee, and
  Hinton]{kornblith2019similarity}
Kornblith, S., Norouzi, M., Lee, H., and Hinton, G.
\newblock Similarity of neural network representations revisited.
\newblock In \emph{International Conference on Machine Learning}, 2019.

\bibitem[LeCun et~al.(1998)LeCun, Bottou, Bengio, Haffner,
  et~al.]{lecun1998gradient}
LeCun, Y., Bottou, L., Bengio, Y., Haffner, P., et~al.
\newblock Gradient-based learning applied to document recognition.
\newblock \emph{Proceedings of the IEEE}, 86\penalty0 (11):\penalty0
  2278--2324, 1998.

\bibitem[Li \& Vasconcelos(2019)Li and Vasconcelos]{REPAIR}
Li, Y. and Vasconcelos, N.
\newblock Repair: Removing representation bias by dataset resampling.
\newblock In \emph{Proceedings of the IEEE Conference on Computer Vision and
  Pattern Recognition}, pp.\  9572--9581, 2019.

\bibitem[Li et~al.(2018)Li, Li, and Vasconcelos]{RESOUND}
Li, Y., Li, Y., and Vasconcelos, N.
\newblock Resound: Towards action recognition without representation bias.
\newblock In \emph{Proceedings of the European Conference on Computer Vision
  (ECCV)}, pp.\  513--528, 2018.

\bibitem[Louppe et~al.(2017)Louppe, Kagan, and Cranmer]{louppe2017learning}
Louppe, G., Kagan, M., and Cranmer, K.
\newblock Learning to pivot with adversarial networks.
\newblock In \emph{Advances in neural information processing systems}, pp.\
  981--990, 2017.

\bibitem[McCoy et~al.(2019)McCoy, Pavlick, and Linzen]{mccoy2019right}
McCoy, R.~T., Pavlick, E., and Linzen, T.
\newblock Right for the wrong reasons: Diagnosing syntactic heuristics in
  natural language inference.
\newblock In \emph{Proceedings of the 57th Annual Meeting of the Association
  for Computational Linguistics}, 2019.

\bibitem[Niven \& Kao(2019)Niven and Kao]{niven2019probing}
Niven, T. and Kao, H.-Y.
\newblock Probing neural network comprehension of natural language arguments.
\newblock In \emph{Proceedings of the 57th Annual Meeting of the Association
  for Computational Linguistics}, 2019.

\bibitem[Panda et~al.(2018)Panda, Zhang, Li, Lee, Lu, and
  Roy-Chowdhury]{panda2018eccv}
Panda, R., Zhang, J., Li, H., Lee, J.-Y., Lu, X., and Roy-Chowdhury, A.~K.
\newblock Contemplating visual emotions: Understanding and overcoming dataset
  bias.
\newblock In \emph{Proceedings of the European Conference on Computer Vision
  (ECCV)}, pp.\  579--595, 2018.

\bibitem[Paszke et~al.(2019)Paszke, Gross, Massa, Lerer, Bradbury, Chanan,
  Killeen, Lin, Gimelshein, Antiga, et~al.]{paszke2019pytorch}
Paszke, A., Gross, S., Massa, F., Lerer, A., Bradbury, J., Chanan, G., Killeen,
  T., Lin, Z., Gimelshein, N., Antiga, L., et~al.
\newblock Pytorch: An imperative style, high-performance deep learning library.
\newblock In \emph{Advances in Neural Information Processing Systems}, pp.\
  8024--8035, 2019.

\bibitem[Peyre et~al.(2017)Peyre, Sivic, Laptev, and Schmid]{peyre2017iccv}
Peyre, J., Sivic, J., Laptev, I., and Schmid, C.
\newblock Weakly-supervised learning of visual relations.
\newblock In \emph{Proceedings of the IEEE International Conference on Computer
  Vision}, pp.\  5179--5188, 2017.

\bibitem[Quadrianto et~al.(2019)Quadrianto, Sharmanska, and
  Thomas]{quadrianto2019discovering}
Quadrianto, N., Sharmanska, V., and Thomas, O.
\newblock Discovering fair representations in the data domain.
\newblock In \emph{Proceedings of the IEEE Conference on Computer Vision and
  Pattern Recognition}, pp.\  8227--8236, 2019.

\bibitem[Ray et~al.(2019)Ray, Sikka, Divakaran, Lee, and
  Burachas]{ray2019sunny}
Ray, A., Sikka, K., Divakaran, A., Lee, S., and Burachas, G.
\newblock Sunny and dark outside?! improving answer consistency in vqa through
  entailed question generation.
\newblock In \emph{Proceedings of the 2019 Conference on Empirical Methods in
  Natural Language Processing and the 9th International Joint Conference on
  Natural Language Processing (EMNLP-IJCNLP)}, pp.\  5860--5865, 2019.

\bibitem[Russakovsky et~al.(2015)Russakovsky, Deng, Su, Krause, Satheesh, Ma,
  Huang, Karpathy, Khosla, Bernstein, et~al.]{russakovsky2015imagenet}
Russakovsky, O., Deng, J., Su, H., Krause, J., Satheesh, S., Ma, S., Huang, Z.,
  Karpathy, A., Khosla, A., Bernstein, M., et~al.
\newblock Imagenet large scale visual recognition challenge.
\newblock \emph{International journal of computer vision}, 115\penalty0
  (3):\penalty0 211--252, 2015.

\bibitem[Sevilla-Lara et~al.(2019)Sevilla-Lara, Zha, Yan, Goswami, Feiszli, and
  Torresani]{sevilla2019onlytimecantell}
Sevilla-Lara, L., Zha, S., Yan, Z., Goswami, V., Feiszli, M., and Torresani, L.
\newblock Only time can tell: Discovering temporal data for temporal modeling.
\newblock \emph{arXiv preprint arXiv:1907.08340}, 2019.

\bibitem[Shah et~al.(2019)Shah, Chen, Rohrbach, and Parikh]{shah2019cycle}
Shah, M., Chen, X., Rohrbach, M., and Parikh, D.
\newblock Cycle-consistency for robust visual question answering.
\newblock In \emph{Proceedings of the IEEE Conference on Computer Vision and
  Pattern Recognition}, pp.\  6649--6658, 2019.

\bibitem[Shetty et~al.(2019)Shetty, Schiele, and Fritz]{shetty2019cvpr}
Shetty, R., Schiele, B., and Fritz, M.
\newblock Not using the car to see the sidewalk--quantifying and controlling
  the effects of context in classification and segmentation.
\newblock In \emph{Proceedings of the IEEE Conference on Computer Vision and
  Pattern Recognition}, pp.\  8218--8226, 2019.

\bibitem[Simonyan \& Zisserman(2015)Simonyan and Zisserman]{simonyan2014very}
Simonyan, K. and Zisserman, A.
\newblock Very deep convolutional networks for large-scale image recognition.
\newblock In \emph{International Conference on Learning Representations}, 2015.

\bibitem[Song et~al.(2012)Song, Smola, Gretton, Bedo, and
  Borgwardt]{unbiasedHSIC}
Song, L., Smola, A., Gretton, A., Bedo, J., and Borgwardt, K.
\newblock Feature selection via dependence maximization.
\newblock \emph{Journal of Machine Learning Research}, 13\penalty0
  (May):\penalty0 1393--1434, 2012.

\bibitem[{Torralba} \& {Efros}(2011){Torralba} and
  {Efros}]{torralba2011unbiased}
{Torralba}, A. and {Efros}, A.~A.
\newblock Unbiased look at dataset bias.
\newblock In \emph{Proceedings of the IEEE Conference on Computer Vision and
  Pattern Recognition}, pp.\  1521--1528, June 2011.

\bibitem[Tran et~al.(2019)Tran, Wang, Torresani, and Feiszli]{tran2019video}
Tran, D., Wang, H., Torresani, L., and Feiszli, M.
\newblock Video classification with channel-separated convolutional networks.
\newblock In \emph{Proceedings of the IEEE International Conference on Computer
  Vision}, pp.\  5552--5561, 2019.

\bibitem[Wang et~al.(2019{\natexlab{a}})Wang, He, and Xing]{HEX}
Wang, H., He, Z., and Xing, E.~P.
\newblock Learning robust representations by projecting superficial statistics
  out.
\newblock In \emph{International Conference on Learning Representations},
  2019{\natexlab{a}}.
\newblock URL \url{https://openreview.net/forum?id=rJEjjoR9K7}.

\bibitem[Wang et~al.(2019{\natexlab{b}})Wang, Zhao, Yatskar, Chang, and
  Ordonez]{wang2019balanced}
Wang, T., Zhao, J., Yatskar, M., Chang, K.-W., and Ordonez, V.
\newblock Balanced datasets are not enough: Estimating and mitigating gender
  bias in deep image representations.
\newblock In \emph{Proceedings of the IEEE International Conference on Computer
  Vision}, pp.\  5310--5319, 2019{\natexlab{b}}.

\bibitem[Weinzaepfel \& Rogez(2019)Weinzaepfel and
  Rogez]{weinzaepfel2019mimetics}
Weinzaepfel, P. and Rogez, G.
\newblock Mimetics: Towards understanding human actions out of context.
\newblock \emph{arXiv preprint arXiv:1912.07249}, 2019.

\bibitem[Zemel et~al.(2013)Zemel, Wu, Swersky, Pitassi, and
  Dwork]{zemel2013icml}
Zemel, R., Wu, Y., Swersky, K., Pitassi, T., and Dwork, C.
\newblock Learning fair representations.
\newblock In \emph{International Conference on Machine Learning}, pp.\
  325--333, 2013.

\bibitem[Zhang et~al.(2018)Zhang, Liu, Liu, Hu, Liu, and Zhu]{zhang2018ijcai}
Zhang, C., Liu, Y., Liu, Y., Hu, Q., Liu, X., and Zhu, P.
\newblock Fish-mml: Fisher-hsic multi-view metric learning.
\newblock In \emph{International Joint Conference on Artificial Intelligence},
  pp.\  3054--3060, 2018.

\end{thebibliography}
